\documentclass[runningheads]{llncs}

\usepackage{eccv}

\usepackage{eccvabbrv}

\usepackage{amsmath,amsfonts,bm}

\def\eqref#1{equation~\ref{#1}}

\def\1{\bm{1}}

\DeclareMathAlphabet{\mathsfit}{\encodingdefault}{\sfdefault}{m}{sl}
\SetMathAlphabet{\mathsfit}{bold}{\encodingdefault}{\sfdefault}{bx}{n}

\DeclareMathOperator*{\argmax}{arg\,max}

\usepackage{graphicx}
\usepackage{booktabs}
\usepackage{twemojis}
\usepackage[table,dvipsnames]{xcolor}    

\usepackage{amsmath}
\usepackage{amssymb}
\usepackage{mathtools}
\usepackage{xspace}
\usepackage{enumitem}
\usepackage[breakable]{tcolorbox}
\usepackage[breaklinks,colorlinks]{hyperref}
\usepackage{xcolor}
\usepackage{wrapfig}
\usepackage[capitalize,noabbrev]{cleveref}

\definecolor{cornflowerblue}{rgb}{0.39, 0.58, 0.93}
\hypersetup{
    colorlinks=true,
    linkcolor=cornflowerblue,
    filecolor=magenta,
    urlcolor=teal,
    citecolor=cornflowerblue,
    pdftitle={A Title},
    pdfpagemode=FullScreen,
}

\usepackage[accsupp]{axessibility}  
\newcommand{\methodshort}{\texttt{Flash}\xspace}
\newcommand{\methodname}{\texttt{Flash-BoN}\xspace}

\usepackage{orcidlink}

\begin{document}

\title{Flash-BoN: Instant Drafts for Inference-Time Scaling in Diffusion Models}


\author{Ruchit Rawal\inst{1} \and
Reza Shirkavand\inst{1} \and
Sayak Paul\inst{2} \and
Yuxin Wen\inst{1} \and
Heng Huang\inst{1} \and
Yizheng Chen\inst{1} \and
Tom Goldstein\inst{1}$^{\star}$ \and
Gowthami Somepalli\inst{1}$^{\star}$}

\authorrunning{R. Rawal et al.}

\institute{
University of Maryland, College Park, MD, USA
\and
Hugging Face, USA
\\[0.5em]
\url{https://flash-bon.github.io}
}

\maketitle
\renewcommand{\thefootnote}{}
\footnotetext{Correspondence to: \email{ruchitr@umd.edu}.\\$^{\star}$~Equal advising.}
\renewcommand{\thefootnote}{\arabic{footnote}}

\begin{abstract}
Inference-time scaling for text-to-image generation has progressed from simple Best-of-$N$ (BoN) sampling to guided search methods that verify and steer candidate trajectories at intermediate denoising steps. These approaches focus on when and how often to verify during denoising but largely treat the cost of generation itself as fixed. Moreover, the standard practice of comparing methods by number of function evaluations (NFEs) counts only denoising forward passes and ignores verifier overhead, which can distort efficiency rankings. We show that under wall-clock evaluation, simple BoN already matches or outperforms several guided search techniques, suggesting that compute is better spent on broader exploration than on repeated intermediate verification. This motivates \methodname{}, which generates a large pool of inexpensive draft candidates by combining three complementary acceleration knobs: timestep truncation, layer skipping, and activation proxies into a single configuration optimized once per model. An efficient multi-stage verification procedure then identifies the most promising draft, which is refined at full quality. Across three benchmarks and three model scales, \methodname{} consistently outperforms all baselines under fixed wall-clock budgets, with gains that grow at larger model scales (+8\% AUC). We further show that our strategy combines well and improves existing orthogonal techniques such as reflection-based prompt optimization (+16\% AUC). The gains correlate with increased candidate diversity, which also enables draft-guided selection to accelerate RL post-training convergence.

\keywords{Inference-time scaling \and Text-to-image generation \and Diffusion models}
\end{abstract}

\begin{figure}[!h]
    \centering
    \includegraphics[width=\linewidth]{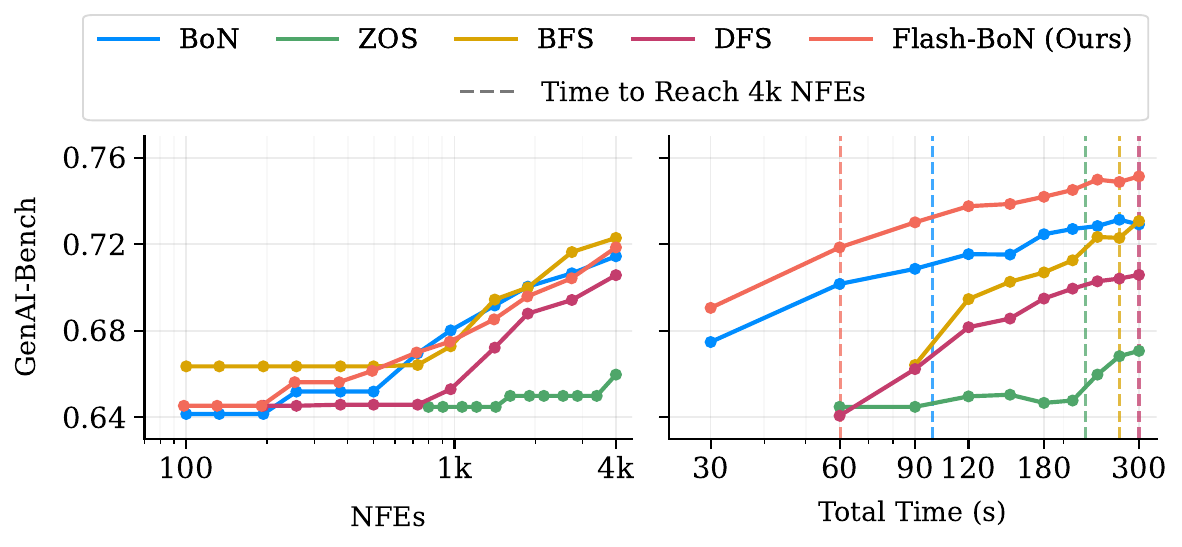}
    \caption{[\textbf{T2I Model:} Wan2.1 1.3B] Comparison under matched NFE and matched wall-clock budgets. \textbf{Left:} At a fixed NFE budget, Breadth-First Search (BFS) is marginally best at high NFEs. \textbf{Right:} Under equal runtime, the trend reverses. Dashed vertical lines mark the wall-clock time each method needs to reach 4k NFEs (the maximum NFE shown on the left). Methods with frequent verifier calls take longer to reach the same NFE budget, reducing exploration within a fixed runtime. Consequently, BoN-style approaches outperform under realistic latency constraints.}
    \label{fig:teaser}
\end{figure}

\section{Introduction}
\label{sec:intro}
Inference-time scaling has become a powerful lever for improving large language models~\cite{wu2025inferencescalinglawsempirical}, where allocating additional compute at test time enables exploration of multiple solution candidates and selection of the best one. Diffusion-based text-to-image (T2I) models have begun to adopt analogous techniques~\cite{Ma_2025_CVPR, singhal2025a, zhang2025inferencetimescalingdiffusionmodels, Zhuo_2025_ICCV}, but existing methods primarily vary \emph{when} and \emph{how often} to verify candidates during intermediate denoising timesteps to assess the potential of ongoing trajectories and prune, guide, or backtrack accordingly. This focus on verification scheduling overlooks an important axis: the cost of generation itself, which we find can be substantially reduced by exploiting redundancies in layer-level computation within and across denoising timesteps.

\looseness=-1
Inference-time scaling for T2I involves two components: \emph{generation}, which produces candidates via the denoising process, and \emph{verification}, which scores them to select the best output. Best-of-$N$ (BoN) sampling sits at one extreme, investing nearly all compute in generation, producing full trajectories from different seeds and invoking the verifier only once at the end. Guided search techniques sit at the other extreme, invoking the verifier at intermediate timesteps to prune or steer trajectories~\cite{zhang2025inferencetimescalingdiffusionmodels, singhal2025a}. This introduces additional design choices such as branching frequency, pruning thresholds, and modifications to the sampling process (e.g., converting deterministic ODE solvers to SDEs). Neither approach questions the per-trajectory generation cost, which raises a natural question: what is the most \textit{efficient way} to allocate a fixed compute budget to verify many potential trajectories and produce the final output for the most promising ones?

Answering this allocation question in practice depends critically on how we measure \emph{compute}. Prior work typically compares inference-time scaling methods using the Number of Function Evaluations (NFEs) as a hardware-agnostic proxy~\cite{Ma_2025_CVPR, singhal2025a, zhang2025inferencetimescalingdiffusionmodels}. However, NFEs only count denoising forward passes and often ignore verifier costs entirely. This matters because the allocation question is fundamentally about \emph{generation versus verification}: if the verifier is unaccounted for, methods that verify more frequently appear efficient even when they are not, especially since the verifier step is often more compute intensive than a single denoising step~\cite{Liu_2025_ICCV}. In fact, once compute is adjusted to account for \emph{all} forward passes, even simple Best-of-$N$ (BoN) can outperform many sophisticated guided techniques. As illustrated in Fig.~\ref{fig:teaser}, BFS appears marginally better under a fixed NFE budget, but this trend reverses under a strict wall-clock budget. Because methods like BFS require frequent intermediate verifier calls, they take substantially longer time to the same NFE limit, sharply reducing the number of candidates they can explore under realistic latency constraints.

\looseness=-1
This motivates \textbf{\methodname}, which combines the strengths of guided search and Best-of-$N$. Instead of spending the budget on repeated intermediate verification, it generates a large pool of inexpensive \emph{draft} candidates, performs a \emph{single} verification step to select the promising ones, and refines only those at full quality.
Our key insight is that several acceleration techniques previously explored in isolation to \emph{reduce generation latency while maintaining quality} can be combined and repurposed for a different objective: \emph{expanding the set of explored candidates under a fixed wall-clock budget}.
Concretely, we combine three complementary acceleration knobs: timestep truncation, layer skipping, and activation proxies into a unified draft-generation configuration that jointly determines the speed fidelity trade-off. We formalize this design as a discrete optimization problem over the configuration space and solve it once per model to find the fastest configuration that preserves sufficient fidelity for verifier preferences to transfer after refinement. The resulting drafts are optimized for wall-clock efficiency by design and require no per-prompt tuning at inference time.

\looseness=-1
Generating a large pool of drafts is only useful if we can reliably identify the one worth refining. Pointwise scoring with an off-the-shelf VLM is efficient but often poorly calibrated, with compressed score ranges that make top candidates hard to distinguish~\cite{whitehouse2505j1, xuan2025seeing}; fully pairwise comparison provides stronger signal but incurs quadratic cost. We introduce a multi-stage selection procedure that combines both: fast pointwise scoring prunes the pool, then targeted pairwise comparisons refine the ranking among a small subset of top candidates, yielding substantially better selections than pointwise alone at a fraction of the pairwise cost.

Together with the multi-stage selection procedure, \methodname forms a complete inference-time scaling pipeline. Across three benchmarks and three model scales, it consistently outperforms all baselines. Our main contributions are:
\vspace{-0.2em}
\begin{enumerate}[leftmargin=*,noitemsep]
    \item \textbf{NFE vs.\ Wall-clock analysis.} NFE comparisons often undercount verifier overhead, distorting method rankings; under wall-clock evaluation, simple BoN already matches or exceeds guided search (Fig.~\ref{fig:teaser}).
    
    \item \textbf{Draft generation and verification.} We show that different acceleration knobs are complementary and formalize their joint configuration as a discrete optimization problem, yielding a draft policy solved once per model. We also introduce a multi-stage verification procedure that pairs fast pointwise pruning with targeted pairwise comparisons to reliably identify the best draft.

    \item \textbf{Consistent gains across scales and benchmarks.} \methodname leads across model--benchmark combinations (Tab.~\ref{tab:main_results}), with margins that grow at larger model scales (+8\% AUC: area under the score-vs-wall-clock curve).

    \item \textbf{Modular combination with orthogonal scaling techniques.} The \methodshort draft strategy functions as a drop-in module that can be layered onto existing methods (such as Reflection-Tuning~\cite{Zhuo_2025_ICCV}), improving AUC by 6--16\%.

    \item \textbf{Diversity analysis \& RL post-training.} Performance gains correlate with increased candidate diversity (Pearson $r{=}0.75$), and the draft-guided selection principle transfers to RL: \methodshort-Flow-GRPO matches baseline convergence in $10\times$ fewer gradient steps.
\end{enumerate}

\section{Related Works}
\label{sec:rel_works}

Recent work has studied allocating additional compute at inference time~\cite{xu2023restart} to improve diffusion model performance. Broadly, existing approaches either \emph{search} for better trajectories~\cite{lee2025adaptivesearch1,he2025scalingsearch2,li2025dynamic,yoon2025monte} or \emph{steer} sampling using a verifier or reward model without retraining the generator~\cite{guo2025trainingfreesteer1,uehara2025inference}. Ma \etal~\cite{Ma_2025_CVPR} formalize \emph{inference-time scaling} as a search over noise space guided by learned verifiers such as ImageReward~\cite{xu2023imagereward}, and show that increasing compute can yield consistent gains. Singhal \etal~\cite{singhal2025a} introduce Feynman--Kac (FK) steering, a particle-based framework that resamples intermediate diffusion states according to reward-defined potentials under deterministic schedulers such as DDIM~\cite{song2021denoising}. Zhang \etal~\cite{zhang2025inferencetimescalingdiffusionmodels} connect inference-time scaling to classical tree search, combining local refinement with global exploration via breadth-first search (BFS) or depth-first search (DFS) over denoising trajectories. Other works extend scaling along additional axes: SANA~1.5~\cite{xie2025sana} studies joint training- and inference-time scaling with best-of-$N$ selection, and ReflectionFlow~\cite{Zhuo_2025_ICCV} performs iterative self-refinement through textual feedback.

Despite their differences, these methods share two characteristics: they measure compute primarily via NFEs or FLOPs, abstracting away verifier and control-logic costs; and they use techniques such as timestep truncation or architectural simplifications to reduce latency while preserving baseline quality. In contrast, we treat \textbf{wall-clock time} as the primary constraint and repurpose these efficiency techniques for a different goal: expanding the set of explored candidates under a fixed runtime budget by combining complementary draft-generation strategies, jointly optimized once per model without repeated intermediate verification.

\section{Problem Setup}
\label{motivation}
Let $\mathcal{D}_\theta$ denote a pre-trained text-to-image diffusion model with parameters $\theta$. Given a text $x \in \mathcal{X}$, standard sampling follows an iterative denoising trajectory through $S$ discrete timesteps $\{t_S, t_{S-1}, \ldots, t_1, t_0\}$, starting from Gaussian noise $z_T \sim \mathcal{N}(0, I)$ and progressively denoising to produce the final image:
$y = \mathcal{D}_\theta(z_T, x, \{t_i\}_{i=1}^S).$
Each denoising step $z_{t_{i-1}} = f_\theta(z_{t_i}, x, t_i)$ applies a neural network $f_\theta$, typically a U-Net~\cite{ronneberger2015unet} or DiT~\cite{peebles2023dit} architecture, consisting of $L$ sequential blocks $\{B_1, B_2, \ldots, B_L\}$. While we describe our method in the context of diffusion models for clarity, it applies equally to flow-matching models, which we also evaluate in our experiments.

Given a prompt $x \in \mathcal{X}$ and a wall-clock time budget $T \in \mathbb{R}^+$, our goal is to return an image $y^{*}$ that maximizes a \emph{verifier} score $V(y,x)$, where $V$ serves as a proxy for a downstream benchmark-specific evaluation metric $E(y,x)$. Inference-time scaling methods therefore generate a pool of candidate images $\{y_j\}_{j=1}^{N}$ and select
\begin{equation}
\label{eq:wall_clock_objective}
y^{*} = \argmax_{y \in \{y_j\}_{j=1}^{N}} V(y, x)
\quad \text{s.t.} \quad
T_{\text{total}}(\{y_j\}_{j=1}^{N}, x) \leq T,
\end{equation}
where $T_{\text{total}}$ encompasses all generation, verification, and (when applicable) data reading/loading costs incurred by the method.

\begin{figure*}[t]
    \centering
    \includegraphics[width=\linewidth]{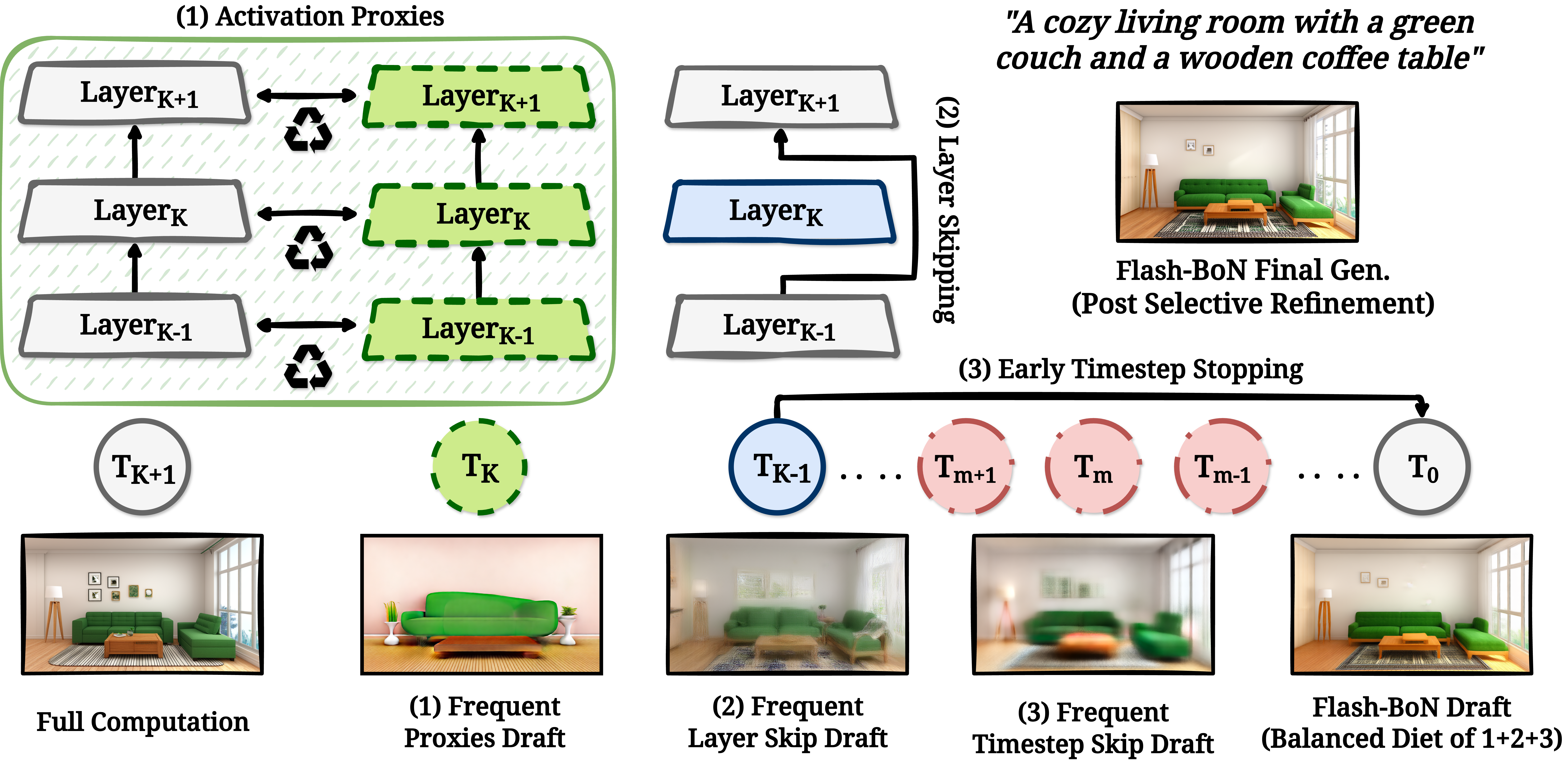}
    \caption{[\textbf{T2I Model:} Wan2.1 1.3B] Acceleration strategies for draft generation: (1) activation proxies exploit redundancies in activations across timesteps, (2) layer skipping omits redundant layers within a timestep, and (3) early stopping truncates the denoising trajectory. The bottom row compares drafts that over-index on each strategy with our \methodname draft (far right), which combines all three through a learned configuration $\phi^*$. The balanced \methodname draft preserves substantially more structure, enabling reliable verification and refinement.}
    \label{fig:methods_fig}
\end{figure*}

\section{\methodname}
\label{method}
\methodname builds on the observation that promising candidates can often be identified without full-cost generation. Rather than relying solely on early timestep truncation as in prior work, we define a draft configuration $\phi$ that combines three acceleration knobs: (i) early timestep stopping, (ii) layer skipping, and (iii) substituting full layer computation with activation proxies. We \emph{precompute} $\phi$ once per model via discrete optimization to obtain $\phi^*$ that is fast while preserving sufficient fidelity. \methodname operates in three stages:
\begin{enumerate}[leftmargin=*,noitemsep]
    \item \textbf{Draft Generation}: Generate $M$ cheap candidates $\{\tilde{y}_1, \ldots, \tilde{y}_M\}$ using $\phi^*$.
    \item \textbf{Verification}: Score all previews with verifier $V$ and select the best draft $\tilde{y}^*$.
    \item \textbf{Refinement}: Resume generation from $\tilde{y}^*$ with full computation to obtain final output $y^*$.
\end{enumerate}

\vspace{-1.5em}
\subsection{Acceleration Knobs \& Precomputing $\phi^*$}
\label{acceleration_knobs}
\vspace{-0.4em}
To generate cheap previews, we must decide which shortcuts to apply without degrading quality. We consider three complementary acceleration techniques:
\vspace{-0.6em}
\paragraph{\textbf{(1) Early Timestep Stopping.}}
Instead of running all $S$ denoising steps, we generate previews using only the first $S' < S$ steps, inspired by ~\cite{li2023autodiffusiontrainingfreeoptimizationtime, wang2025pfdiff}. Let $z_t \in \mathbb{R}^{d}$ denote the latent at timestep $t$. The full denoising trajectory is:
\begin{equation}
z_{t_{i-1}} = f_\theta(z_{t_i}, x, t_i) \quad \text{for } i = S, S-1, \ldots, 1,
\end{equation}
where $f_\theta$ is the denoising network. Early stopping truncates this to $i = S, \ldots, S - S' + 1$.

\paragraph{\textbf{(2) Activation Proxies.}}
We can reduce computation by using \emph{activation proxies} computed from cached features at previously visited timesteps~\cite{chen2024deltadittrainingfreeaccelerationmethod,habibian2024clockworkdiffusionefficientgeneration,agarwal2023approximatecachingefficientlyserving}. Since denoising proceeds from high to low timesteps ($t_S \to t_0$), when processing step $i$ the activations at $t_{i+1}, t_{i+2}, \ldots$ are already available. Following Liu \etal~\cite{Liu_2025_ICCV}, we forecast the activation $h_{t_i}^{(\ell)}$ for layer $\ell$ via a second-order Taylor expansion anchored at these cached values. Let $\Delta_t = t_i - t_{i+1}$. We estimate temporal derivatives from cached features:
\begin{equation}
\dot{h}_{t_{i+1}}^{(\ell)} \approx 
\frac{h_{t_{i+1}}^{(\ell)} - h_{t_{i+2}}^{(\ell)}}{t_{i+1} - t_{i+2}},
\end{equation}
and
\begin{equation}
\ddot{h}_{t_{i+1}}^{(\ell)} \approx 
\frac{h_{t_{i+1}}^{(\ell)} - 2h_{t_{i+2}}^{(\ell)} + h_{t_{i+3}}^{(\ell)}}{(t_{i+1} - t_{i+2})^2},
\end{equation}
and then form the activation proxy:
\begin{equation}
\tilde{h}_{t_i}^{(\ell)} =
h_{t_{i+1}}^{(\ell)} + \Delta_t \dot{h}_{t_{i+1}}^{(\ell)} + \frac{1}{2}\Delta_t^{2} \ddot{h}_{t_{i+1}}^{(\ell)},
\end{equation}
which is used as the layer output in place of a full forward computation.

\vspace{-1em}
\paragraph{\textbf{(3) Layer Skipping.}}
While activation proxies exploit redundancy \emph{across timesteps} within a layer, we can also leverage redundancy \emph{across layers} within a single timestep \cite{zhu2024dipgodiffusionprunerfewstep,chen2025stabilizedefficientdiffusiontransformers}. To do this, we selectively skip a subset of layers $\mathcal{L}_{\text{skip}} \subset \{1, \ldots, L\}$ during denoising. At timesteps where skipping is applied, if layer $\ell$ is skipped, we simply propagate the previous activation:
\begin{equation}
h^{(\ell)}_{t_i} = h^{(\ell-1)}_{t_i}
\end{equation}
instead of computing the full block output:
\begin{equation}
h^{(\ell)}_{t_i} = B_\ell(h^{(\ell-1)}_{t_i}, x, t_i),
\end{equation}
where $B_\ell$ denotes the $\ell$-th block of the denoising network.

\noindent \textbf{Configuration Space.}
The three knobs above define a joint configuration $\phi = (S', n_{\text{skip}}, \ell_{\text{start}}, \ell_{\text{end}}, f_{\text{full}})$, where $S'$ is the number of denoising steps for previews, $n_{\text{skip}}$ is the number of initial timesteps before layer skipping begins, $\ell_{\text{start}}, \ell_{\text{end}} \in [1, L]$ specify the contiguous range of layers to skip that is $\mathcal{L}_{\text{skip}} = \{\ell_{\text{start}}, \ldots, \ell_{\text{end}}\}$, and $f_{\text{full}}$ controls how frequently full (non-proxied) computation is performed. Each $\phi$ induces a different speed--fidelity trade-off for draft generation; Fig.~\ref{fig:methods_fig} illustrates how over-indexing on any single knob degrades quality compared to a balanced combination at the same speedup.

\noindent \textbf{Precomputing $\phi^*$.}
We select an optimal configuration $\phi^*$ once per model by solving a black-box discrete optimization problem that jointly maximizes wall-clock speedup and perceptual similarity (LPIPS) to full-compute outputs, evaluated on a held-out calibration set of 120 prompts. We use dual annealing~\cite{Tsallis:1987eu, XIANG1997216} as the optimizer and select $\phi^*$ from the resulting Pareto frontier by maximizing speedup subject to a minimum similarity constraint. The optimization runs for 200 iterations; the resulting $\phi^*$ is then reused across all prompts and benchmarks at inference time. This step is important as the same knobs in a random configuration produce drafts too degraded for the verifier to rank, falling below BoN (App.~\ref{appendix_subsec:component_ablations}). Full details of the objective, optimizer comparisons, calibration set design, and the Pareto frontier are provided in Appendix~\ref{appendix_subsec:addn_optimization_details}.

\vspace{-1em}
\subsection{\methodname Stages}
\label{stages}
\paragraph{\textbf{Stage 1: Draft Generation.}}

Using the precomputed configuration $\phi^*$, we generate a pool of $M$ preview candidates $\{\tilde{y}_1, \ldots, \tilde{y}_M\}$ by sampling different random seeds. Each preview is generated as:
\begin{equation}
\tilde{y}_j = \mathcal{D}_\theta(z_T^{(j)}, x, \phi^*), \quad z_T^{(j)} \sim \mathcal{N}(0, I),
\end{equation}
where $\mathcal{D}_\theta(\cdot, \cdot, \phi^*)$ denotes the model with acceleration configuration $\phi^*$ applied.

\paragraph{\textbf{Stage 2: Verification and Selection.}}

Given $M$ draft candidates $\{\tilde{y}_j\}_{j=1}^{M}$, the simplest selection strategy is \emph{pointwise scoring}: apply an off-the-shelf verifier $V_{\text{pw}}$ independently to each candidate and retain the highest-scoring one,
\begin{equation}
j^* = \argmax_{j \in [M]} \; s_j, 
\qquad 
s_j = V_{\text{pw}}(\tilde{y}_j, x) \in \{1,\ldots,10\}.
\end{equation}
\looseness=-1
This is cheap and fully parallelizable, but often unreliable: VLM-based verifiers are noisy and poorly calibrated as absolute scorers~\cite{whitehouse2505j1}, and integer-valued outputs compress resolution at the top ranks. In practice, over 80\% of prompts produce tied highest scores, with roughly 30\% having 20+ candidates sharing the same top score, making selection effectively random (Appendix~\ref{appendix_subsec:pairwise_vs_pointwise}). \emph{Pairwise} comparison, where the verifier judges two candidates head-to-head, avoids score compression and is better calibrated for relative quality, but incurs prohibitive $\mathcal{O}(M^2)$ cost when applied exhaustively.

\textbf{Multi-stage selection with Elo ranking.}
We combine both views in a progressive filter that maintains Elo ratings~\cite{Elo1978TheRO} throughout. Let $\rho_k \in (0,1]$ be the retention ratio at stage $k$ and $\kappa \ge 0$ the average sparse pairwise comparisons per survivor.

\begin{enumerate}[leftmargin=*,noitemsep]
    \item \textbf{Pointwise pruning + Elo seeding.} Score all $M$ candidates and retain the top-$\rho_1$ fraction. Each candidate's Elo rating is initialized from its pointwise score via $r_j^{(0)} = 1500 + (s_j - 5)\cdot 100$, seeding subsequent stages with an informed prior:
    \begin{equation}
    \mathcal{S}_1 = \mathrm{Top}_{\lceil \rho_1 M \rceil}\!\bigl(\{r_j^{(0)}\}_{j=1}^M\bigr), \quad M_1 \coloneqq |\mathcal{S}_1|.
    \end{equation}

    \item \textbf{Sparse pairwise (adjacent comparisons).} Sort $\mathcal{S}_1$ by Elo and form $C_2 = \lceil \kappa M_1 \rceil$ pairs between \emph{adjacent} entries, concentrating comparisons where the ranking is most uncertain. Each outcome updates both ratings via Elo with a confidence-weighted $K$-factor.\footnote{The pairwise verifier returns a confidence level (high/medium/low), scaling $K$ by 1.2/1.0/0.7 respectively.} Retain the top-$\rho_2$ fraction:
    \begin{equation}
    \mathcal{S}_2 = \mathrm{Top}_{\lceil \rho_2 M_1 \rceil}\!\bigl(\{r_j\}_{j \in \mathcal{S}_1}\bigr), \quad M_2 \coloneqq |\mathcal{S}_2|.
    \end{equation}

    \item \textbf{Dense pairwise tournament.} Run all $\binom{M_2}{2}$ comparisons within $\mathcal{S}_2$, update ratings, and select:
    \begin{equation}
    \tilde{y}^{*} = \tilde{y}_{j^*}, \qquad j^* = \argmax_{j \in \mathcal{S}_2}\; r_j.
    \end{equation}
\end{enumerate}

With our defaults ($\rho_1 {=} \rho_2 {=} 0.5$, $\kappa {=} 1.5$), the procedure uses roughly $\tfrac{3}{2}M$ verifier calls compared to the $\tfrac{1}{2}M^2$ required by exhaustive pairwise ranking. We ablate alternative ranking algorithms (Swiss tournaments, knockout brackets, and full pairwise Elo), and effect of varying $\rho_1, \rho_2, \kappa$ in Appendix~\ref{appendix_subsec:pairwise_vs_pointwise}.

We distinguish the \emph{verifier} $V$ used at inference time from the \emph{evaluation metric} $E$ used for reporting. The verifier guides candidate selection and is typically an off-the-shelf VLM or reward model, while $E$ denotes benchmark-facing metrics such as programmatic checkers or evaluation-tuned VLMs. Using the same model for both conflates selection with evaluation and risks metric overfitting~\cite{Ma_2025_CVPR}. We provide qualitative examples for this in Fig.~\ref{fig:verifier_eval_disagreement}.

\paragraph{\textbf{Stage 3: Selective Refinement}.}

After selecting the best candidate $\tilde{y}^*$, we resume generation from its cached latent state and perform the remaining denoising steps with full computation (i.e., no skipping or proxying). Let $S'$ denote the number of steps used during draft generation. We complete the rest of the trajectory as:
\begin{equation}
y^* = \mathcal{D}_\theta(z_{t_{S'}}^*, x, \phi_{\text{full}}),
\end{equation}
where $z_{t_{S'}}^*$ is the latent at the draft truncation point and $\phi_{\text{full}}$ corresponds to the standard, unaccelerated configuration.
This is valid because the diffusion sampling process is Markovian: future outputs depend only on the current latent, not on how it was produced. The cached state comprises the latent tensor $z_{t_{S'}}$, optionally the layer activations $\{h^{(\ell)}_{t_{S'}}\}_{\ell=1}^L$ for warm-starting activation proxies, and the random seed for reproducibility. To limit GPU memory usage during draft generation, we offload cached tensors to CPU and transfer only the selected candidate back for refinement. This data-transfer overhead is absent from NFE accounting but fully captured by our wall-clock evaluation; despite it, \methodname still outperforms BoN and BFS in end-to-end runtime.

\section{Experiments}
\label{sec:experiments}
We compare \methodname against four baselines: Best-of-$N$ (BoN), Breadth-First Search (BFS), Depth-First Search (DFS)~\cite{zhang2025inferencetimescalingdiffusionmodels}, and Zeroth-Order Search (ZOS)~\cite{Ma_2025_CVPR}. Together these cover the main families of inference-time scaling, from simple global exploration to guided trajectory search. For flow-based models (FLUX.1-dev), which use deterministic ODE solvers, we convert to an SDE to enable trajectory branching (Appendix~\ref{appendix_subsec:ode_sde}). We evaluate on GenAI-Bench~\cite{li2024genai} (300 prompts), GenEval~\cite{ghosh2023geneval} (full), and UniGenBench~\cite{wang2025pref} (300 prompts). For GenAI-Bench and UniGenBench, we uniformly subsample from the evaluation set, as running all methods for 300\,s per prompt on the full benchmarks would be computationally prohibitive (details in Appendix~\ref{appendix_subsec:eval_protocol_details}). Experiments span three model scales: Wan 2.1 1.3B, Wan 2.1 14B~\cite{wan2025wanopenadvancedlargescale}, and FLUX.1-dev~\cite{flux1dev2024}. All methods run on NVIDIA H200 GPUs with end-to-end wall-clock timing covering generation, verification, refinement, and data transfer\footnote{We incorporated warming up of the system to reduce variance.}.

\begin{figure}
    \centering
    \includegraphics[width=\linewidth]{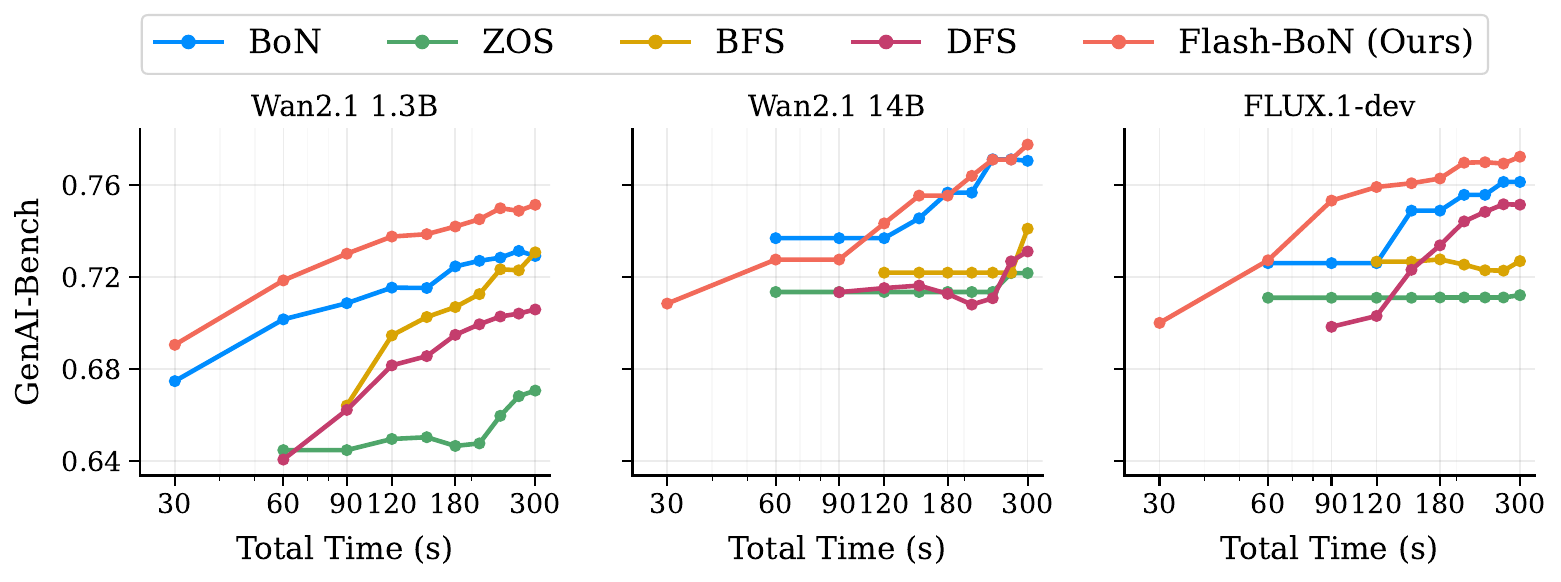}
    \caption{\textbf{GenAI-Bench under fixed wall-clock budgets.} 
    Performance versus total runtime (seconds). All methods are evaluated under identical hardware and end-to-end time accounting. Best-of-$N$ (BoN) consistently outperforms guided search baselines across models. \methodname achieves the strongest performance at all time budgets, with gains that persist and widen at larger scales.}
    \label{fig:genai_time}
\end{figure}

\paragraph{\textbf{Performance under wall-clock constraints.}}
Fig.~\ref{fig:genai_time} extends the single-model comparison from Fig.~\ref{fig:teaser} to three architectures spanning an order of magnitude in parameter count: Wan2.1 1.3B, Wan2.1 14B, and FLUX.1-dev. Across all three models, BoN-style approaches that prioritize broad exploration consistently outperform guided search methods that invest compute in intermediate verification. BFS, DFS, and ZOS produce modest gains on the smaller 1.3B model but scale poorly to larger architectures; on Wan2.1 14B and FLUX.1-dev their curves are nearly flat, indicating that repeated verifier calls leave little room for additional candidates as per-image cost grows. \methodname improves upon BoN across nearly the entire runtime range, with gains that widen as the budget increases and persist across model sizes. Table~\ref{tab:main_results} summarizes these trends via AUC/Time, the area under the score-versus-wall-clock curve normalized by total time, and shows that \methodname leads in every model--benchmark combination across GenAI-Bench, GenEval, and UniGenBench. The advantage over the next best baseline, BoN, is most pronounced on larger models (Wan2.1 14B and FLUX.1-dev), where higher per-image generation cost makes inexpensive drafts particularly effective for expanding exploration within a fixed budget. A practical implication is that Wan2.1 1.3B with \methodname at 300\,s reaches GenAI-Bench scores comparable to Wan2.1 14B at 120--150\,s, suggesting that inference-time scaling can partially substitute for model size under deployment constraints.  Additionally, we break down each knob's contribution in App.~\ref{appendix_subsec:component_ablations}, and show in App.~\ref{appendix_subsec:few_step} that the draft-and-select approach still pays off on few-step distilled models like FLUX.1-schnell, where the per-step budget is already small.
\begin{table}[t]
\centering
\caption{Normalized AUC ($\uparrow$) comparison of inference-time search techniques across benchmarks and models. Best results per column are \textbf{bold}.}
\label{tab:main_results}
\resizebox{0.9\textwidth}{!}{%
\begin{tabular}{l ccc ccc ccc}
\toprule
 & \multicolumn{3}{c}{GenAI-Bench} & \multicolumn{3}{c}{GenEval} & \multicolumn{3}{c}{UniGenBench} \\
\cmidrule(lr){2-4} \cmidrule(lr){5-7} \cmidrule(lr){8-10}
Method & Wan 1.3B & Wan 14B & FLUX & Wan 1.3B & Wan 14B & FLUX & Wan 1.3B & Wan 14B & FLUX \\
\midrule
ZOS & 0.52 & 0.57 & 0.57 & 0.29 & 0.36 & 0.37 & 0.45 & 0.52 & 0.49 \\
DFS & 0.55 & 0.50 & 0.51 & 0.36 & 0.37 & 0.35 & 0.47 & 0.46 & 0.45 \\
BFS & 0.50 & 0.43 & 0.44 & 0.31 & 0.28 & 0.29 & 0.43 & 0.39 & 0.38 \\
BoN & \underline{0.65} & \underline{0.60} & \underline{0.60} & \underline{0.43} & \underline{0.42} & \underline{0.42} & \underline{0.54} & \underline{0.53} & \underline{0.54} \\
\midrule
\rowcolor{gray!20}
\textsc{\methodname} & \textbf{0.66} & \textbf{0.68} & \textbf{0.68} & \textbf{0.45} & \textbf{0.49} & \textbf{0.47} & \textbf{0.57} & \textbf{0.60} & \textbf{0.61} \\
\bottomrule
\end{tabular}
}
\end{table}

\begin{figure}
    \centering
    \includegraphics[width=\linewidth]{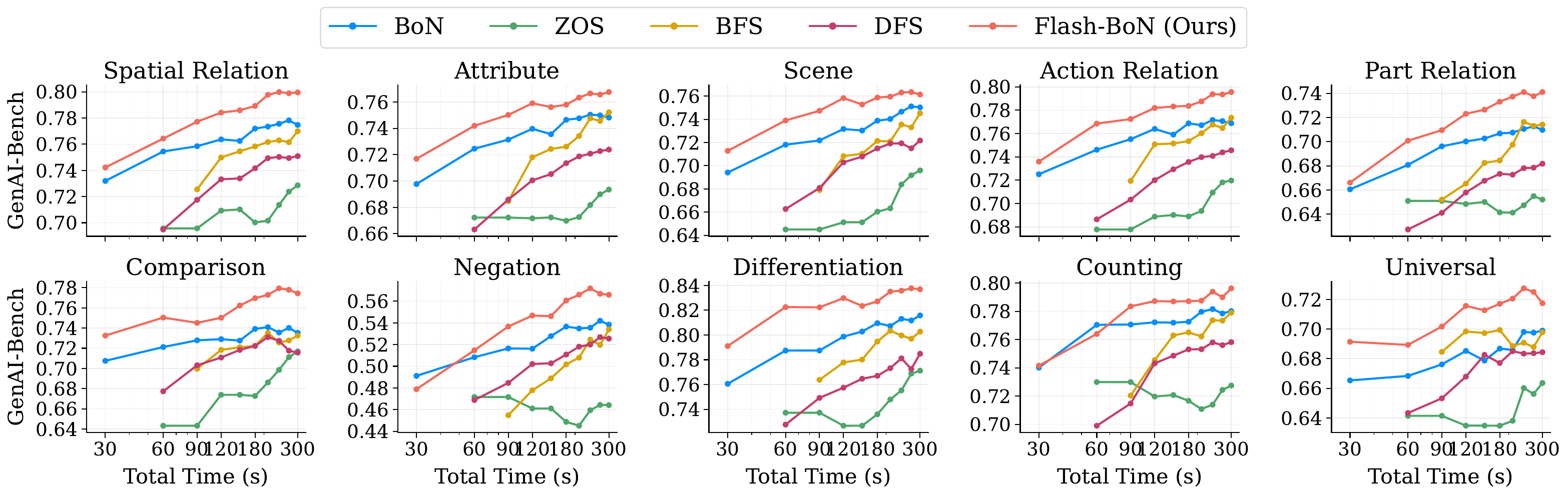}
    \caption{Per-category GenAI-Bench scores on Wan2.1 1.3B under wall-clock time budgets. \methodname leads across all ten categories.}
    \label{fig:genai_categories}
\end{figure}

\begin{figure}[!ht]
    \centering
    \includegraphics[width=\linewidth]{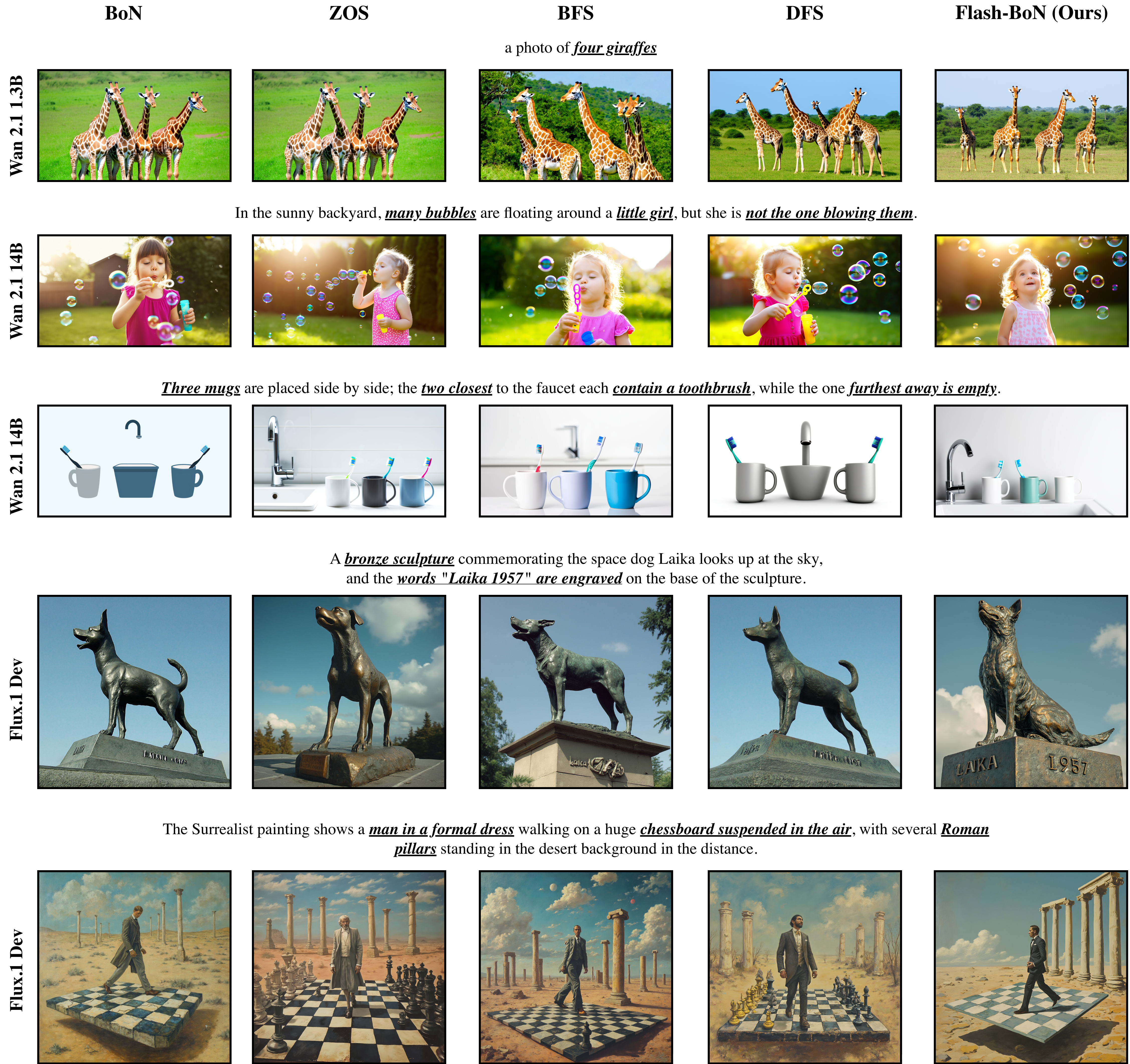}
    \caption{\textbf{Qualitative comparison across methods.} Each row shows the best image selected by each technique under a fixed 300\,s wall-clock budget. Prompts span spatial relations, negation, counting, and complex scene composition. \methodname consistently produces images that more faithfully satisfy the prompt.}
    \label{fig:qualitative_comparison}
\end{figure}

\paragraph{\textbf{Performance across different prompt categories.}}
Fig.~\ref{fig:genai_categories} breaks down performance on Wan2.1 1.3B by GenAI-Bench prompt category. \methodname leads in every category, with the largest gains on compositional prompts (Spatial Relation, Comparison, Part Relation), where the primary failure mode is incorrect layout or missing structure; a broader draft pool increases the chance of sampling a valid composition. The gap narrows for fine-grained categories such as Attribute and Counting, where relevant distinctions are harder to resolve from draft-quality previews. Among guided search methods, BFS is most competitive on Counting and Attribute, where partially formed images already carry informative signal. On holistic categories such as Universal and Negation, where correctness depends on full image context, intermediate verification provides less actionable information and the overhead dominates. Fig.~\ref{fig:qualitative_comparison} compares the best candidates selected by each method across a range of prompts.

\begin{figure}[t]
    \centering
    \begin{minipage}[t]{0.48\linewidth}
        \centering
        \includegraphics[width=\linewidth]{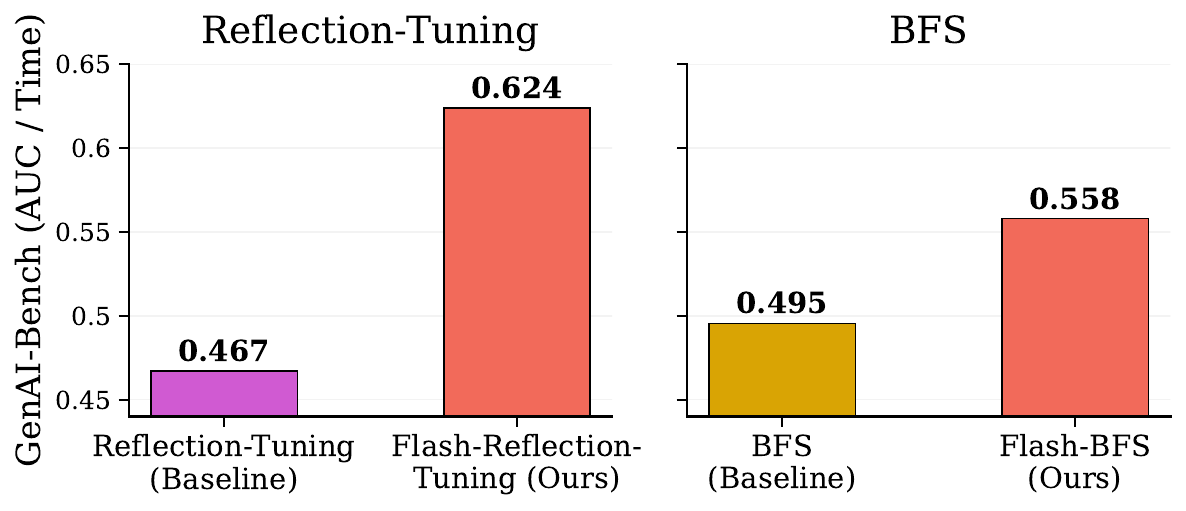}
        \captionof{figure}{\textbf{Combining the \methodshort draft strategy with existing inference-time scaling methods on Wan2.1 1.3B (GenAI-Bench)}. We report AUC/Time, the area under the score-vs-wall-clock curve normalized by total time, as a summary efficiency metric. Applying \methodshort drafts to Reflection-Tuning yields a large improvement (0.46 to 0.62), while the gain on BFS is more modest (0.49 to 0.55).}
        \label{fig:composability}
    \end{minipage}%
    \hfill
    \begin{minipage}[t]{0.48\linewidth}
        \centering
        \includegraphics[width=\linewidth]{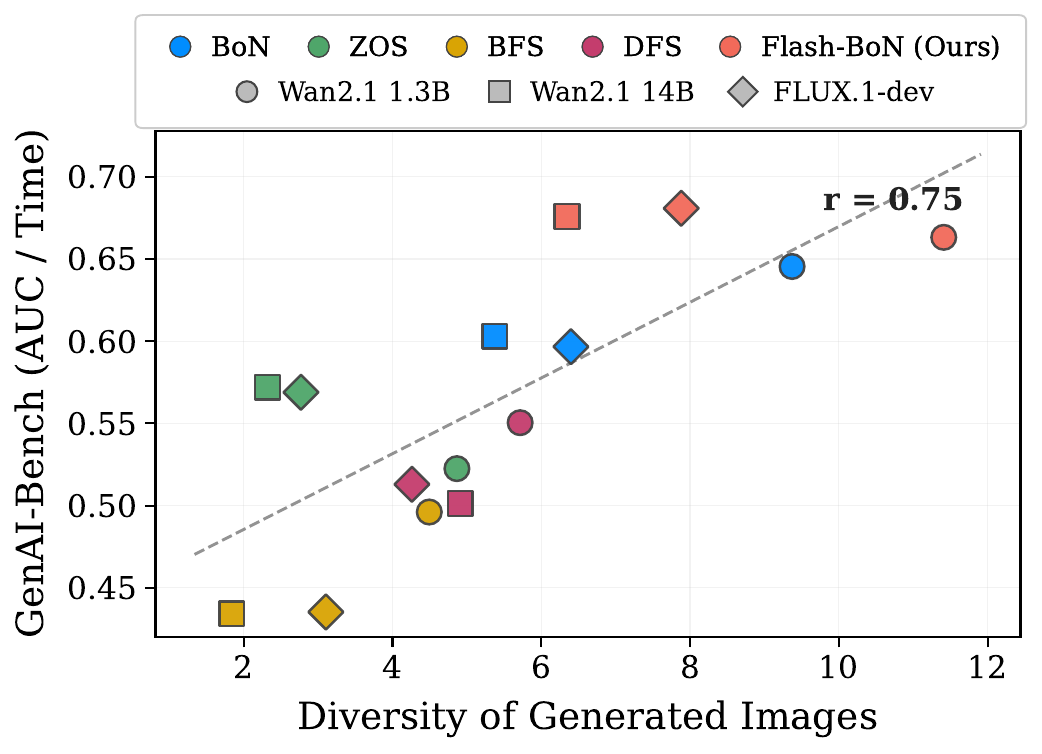}
        \captionof{figure}{\textbf{Candidate pool diversity vs.\ final GenAI-Bench performance across three models}. Diversity is measured using the Vendi Score.
        Each point is one method averaged over all prompts; color indicates technique, shape indicates model.
        Dashed line shows the linear correlation fit.
        Methods that explore more diverse candidates consistently achieve higher quality (Pearson $r{=}0.75$).}
        \label{fig:diversity_analysis}
    \end{minipage}
\end{figure}

\paragraph{\textbf{Combining \methodshort drafts with existing scaling methods.}}
The draft strategy can be layered onto existing inference-time scaling algorithms (Fig.~\ref{fig:composability}). We demonstrate this on two representative methods: Reflection-Tuning~\cite{Zhuo_2025_ICCV} and BFS~\cite{zhang2025inferencetimescalingdiffusionmodels}. Reflection-Tuning iteratively generates an image, critiques it via a VLM, and revises the text prompt accordingly; each cycle requires a full image generation. In \methodshort-Reflection-Tuning, the critique loop operates on draft images instead, and a full-quality generation is produced only after the prompt converges, substantially reducing the cost per cycle. BFS prunes trajectories via intermediate verification; in \methodshort-BFS, we accelerate the steps between verification points using drafts, reverting to full computation only after the final pruning decision (see Appendix~\ref{appendix_subsec:composability_details} for implementation details). To summarize performance across all time budgets in a single number, we report AUC/Time, the area under the score-versus-wall-clock curve normalized by total time.

\looseness=-1
\methodshort-Reflection-Tuning improves AUC from 0.46 to 0.62, a large gain because cheaper cycles enable more prompt revisions within the same budget. \methodshort-BFS improves more modestly, from 0.49 to 0.55: faster intermediate steps allow more candidates, but BFS searches locally and cannot escape to distant regions of noise space where compositionally correct solutions may lie. The contrast is instructive: methods that expand the effective search space benefit most from cheaper drafts, while those that refine within a fixed region see diminishing returns.

\paragraph{\textbf{Why does broader exploration help? A diversity analysis.}}
The results above show that exploring more seeds under a fixed budget consistently improves quality. A natural question is whether this simply reflects more lottery tickets or whether effective methods cover a genuinely more \emph{diverse} region of the output space. To examine this, we measure the diversity of each method's candidate pool using the \emph{Vendi Score}~\cite{friedman2023vendi}, defined as $\mathrm{VS}(K) = \exp\!\bigl(-\sum_{i} \hat{\lambda}_i \log \hat{\lambda}_i\bigr)$ with $\hat{\lambda}_i = \lambda_i / \mathrm{tr}(K)$, where $K$ is a pairwise similarity matrix computed from DINOv2~\cite{oquab2023dinov2} features. Intuitively, VS equals 1 for $n$ identical images and $n$ for $n$ perfectly distinct ones. We compute it over all candidates generated within the 300\,s budget for each method--model--prompt combination.
As shown in Fig.~\ref{fig:diversity_analysis}, diversity and final quality are positively correlated (Pearson $r{=}0.75$). Guided search methods cluster in the low-diversity, lower-performance region, while \methodname consistently sits at the high-diversity end, suggesting that inexpensive drafts enable not only more candidates but also \emph{meaningfully different} ones (qualitative examples in Appendix~\ref{appendix_subsec:diversity_qualitative}).

\begin{figure}[t]
    \centering
    \begin{minipage}[t]{0.48\linewidth}
        \centering
        \includegraphics[width=\linewidth]{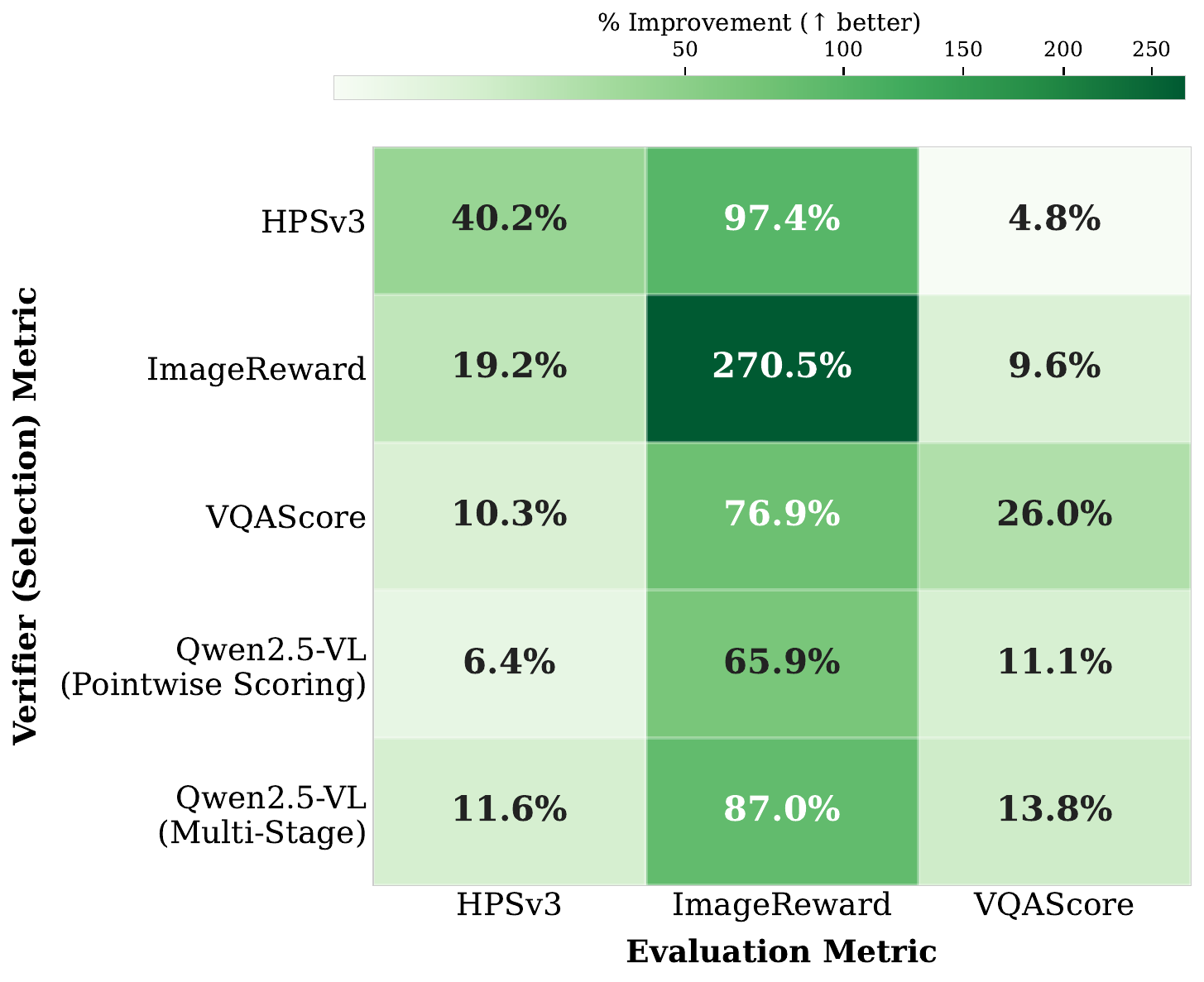}
        \captionof{figure}{\textbf{Percent improvement over no-scaling baseline (Wan 2.1 1.3B).}
        Each cell reports relative improvement when using a given verifier (rows) and evaluating with a specific metric (columns). Diagonal entries, where verifier matches evaluation metric, show the largest gains, while off-diagonal entries reflect cross-metric generalization.}
        \label{fig:verifier_heatmap}
    \end{minipage}%
    \hfill
    \begin{minipage}[t]{0.48\linewidth}
        \centering
        \includegraphics[width=\linewidth]{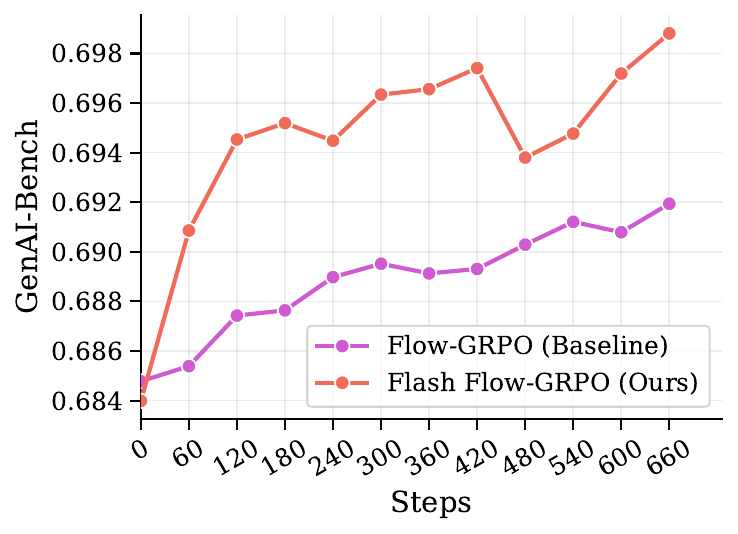}
        \captionof{figure}{\textbf{RL post-training with \methodname rollouts.} GenAI-Bench (VQAScore) during Flow-GRPO training on Wan2.1 1.3B. \methodshort-Flow-GRPO converges faster and maintains a consistent advantage throughout training.}
        \label{fig:grpo_run}
    \end{minipage}
\end{figure}

\paragraph{\textbf{Effect of verifier and evaluation metric.}}
Next, we examine how inference-time scaling interacts with the choice of verifier and eval.\ metric. For each verifier $V$ and metric $E$, we report the percent improvement of \methodname over the no-scaling baseline (Fig.~\ref{fig:verifier_heatmap}). Gains are largest along the diagonal, where verifier matches evaluator, but the magnitude varies strikingly: ImageReward shows 270\% improvement versus 40\% for HPSv3 and 26\% for VQAScore. We hypothesize that ImageReward is particularly susceptible to reward over-optimization: its no-scaling baseline is low relative to the post-scaling score, yet once even a small candidate pool is available, scores plateau almost immediately and remain flat (Appendix~\ref{appendix_subsec:score_trajectories}), suggesting the model locks onto a narrow preference and stops discriminating. This is consistent with findings that holistic preference models tend to favor aesthetic outputs even when they deviate from the prompt~\cite{guo2025aesthetic, xu2023imagereward}. We share qualitative examples of this failure mode in Appendix~\ref{appendix_subsec:verifier_eval_diff}.

\begin{figure}[!t]
    \centering
    \includegraphics[width=\linewidth]{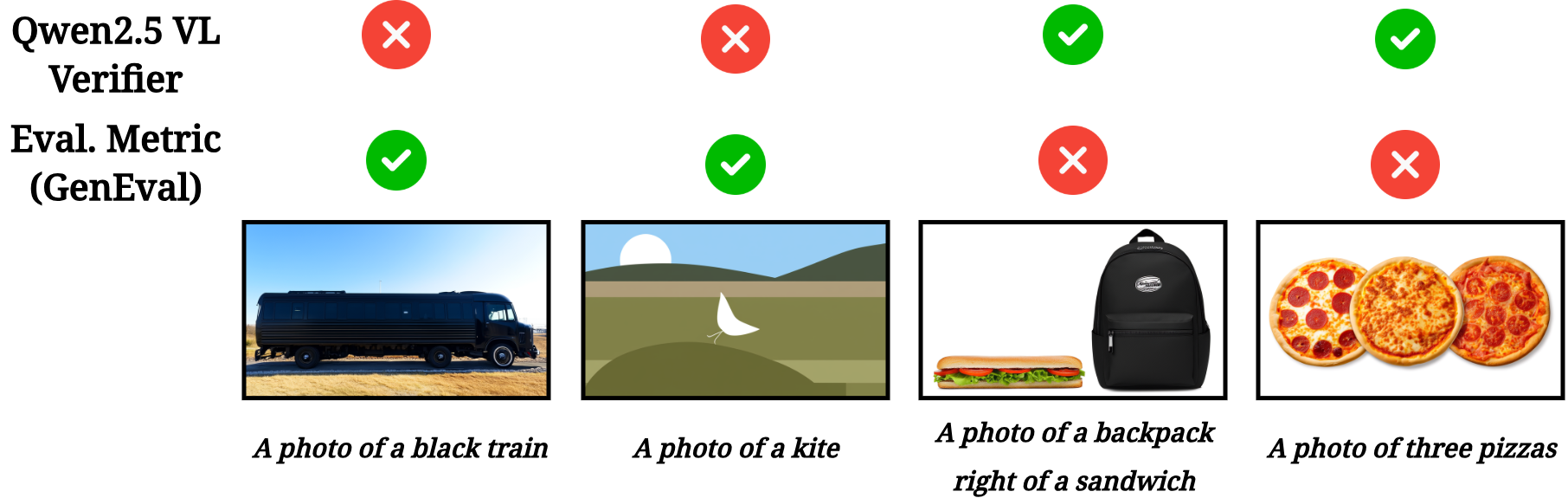}
    \caption{\textbf{Failure modes when verifier and evaluator coincide (Wan 2.1 1.3B).} Left two columns: the evaluator incorrectly accepts flawed outputs (false positives). Right two columns: the evaluator incorrectly rejects valid outputs (false negatives). When the same model serves both roles, inference-time scaling may exploit these biases, influencing scores without genuine quality improvement.}
    \label{fig:verifier_eval_disagreement}
\end{figure}

More broadly, no evaluation metric is bias-free: learned preference models are constrained by their training data and annotation protocols~\cite{ma2508hpsv3, xu2023imagereward}, and optimizing against any single metric risks exploiting its blind spots, producing artifacts not present in the prompt~\cite{ma2508hpsv3}. Fig.~\ref{fig:verifier_eval_disagreement} illustrates this: when the same model acts as both verifier and evaluator, scaling selects for that model's biases rather than true quality. A distinct proxy verifier avoids this loop: in our setup, general VLM verifiers produce consistent gains across \emph{all} evaluation metrics (6--14\%), even without topping any single matched column, indicating stronger cross-metric generalization. Our multi-stage verification further improves over pointwise scoring across all three metrics, supporting our selection design.

\vspace{-1em}
\paragraph{\textbf{Application to RL post-training.}}
\looseness=-1
The draft-and-select strategy also applies beyond inference-time scaling. In RL post-training methods such as Flow-GRPO~\cite{liu2025flow}, the model generates a group of $G$ rollout trajectories per prompt, scores them with a reward model, and updates the policy via group-relative advantage estimation. Sampling these rollouts is both memory and time intensive, so trajectories are typically drawn at random. For difficult prompts, this makes it unlikely that any rollout in a small group will produce a strong positive example, leaving the optimizer with limited signal. In \methodshort-Flow-GRPO, we replace random sampling with draft-guided selection: generate $2G{=}16$ inexpensive draft rollouts, score them, and retain the top 6 and bottom 2 candidates for gradient computation, performing full forward passes only on these $G{=}8$ selected trajectories. The broader draft pool increases the probability of encountering informative positive and negative examples. All hyperparameters are identical to the baseline except for rollout selection and a slightly higher KL penalty to account for the draft-selected distribution (Appendix~\ref{appendix_subsec:grpo_details}).

Fig.~\ref{fig:grpo_run} shows results on Wan2.1 1.3B. \methodshort-Flow-GRPO opens a clear gap within the first 60 steps and maintains it throughout training, reaching 0.699 vs.\ 0.692 for the baseline. Notably, \methodshort-Flow-GRPO at step 60 already matches the baseline's final performance, suggesting roughly $10\times$ faster convergence. We show qualitative examples for the same in Appendix~\ref{appendix_sec:qual_example_grpo}.

\section{Discussion}
\looseness=-1
We introduced \methodname{}, an inference-time scaling method that generates a large pool of inexpensive draft candidates, identifies the most promising one via an efficient multi-stage verification procedure, and refines only that candidate at full quality. Across multiple benchmarks and models, \methodname{} consistently outperforms baselines under fixed wall-clock budgets, with gains that grow at larger model scales. These improvements are driven by genuinely broader and more diverse exploration (Pearson $r{=}0.75$), combine well with orthogonal techniques such as Reflection-Tuning, and transfer to RL post-training, where \methodshort-Flow-GRPO matches baseline convergence while requiring $10\times$ fewer gradient steps.
While our results show strong improvements, several open questions remain for future work, such as, extending draft-and-refine scaling to text-to-video with policies that retain temporal consistency and verifiers trained specifically to compare draft-quality images to improve reliability.

\section*{Acknowledgments}
This work was supported by DOE Office of Science’s ASCR AI for Science initiative, the NSF TRAILS Institute (2229885), and Coefficient Giving, and Longview Philanthropy. R.S. and H.H. were partially supported by NSF IIS 2347592, 2348169, DBI 2405416, CCF 2348306, CNS 2347617, RISE 2536663.

\clearpage
\bibliographystyle{splncs04}
\bibliography{main}


\clearpage
\appendix

\centering

{\large \textbf{Flash-BoN: Instant Drafts for Inference-Time Scaling in Diffusion Models}} \\
\vspace{2.0em}{\large Supplemental Material} \\
\vspace{1.5em}
\raggedright

\section{Additional Experiment Setup Details}
\label{appendix_sec:addn_experiment_setup}

\subsection{Enabling Guided Search on Flow Models via ODE-to-SDE Conversion}
\label{appendix_subsec:ode_sde}
Guided search baselines (BFS, DFS, ZOS) rely on branching or resampling at intermediate denoising steps: from a shared latent state, multiple child trajectories are spawned by drawing different noise realizations. This mechanism requires stochasticity in the sampling process. Diffusion models support this natively, since the reverse process is an SDE with explicit noise injection at every step. Flow matching models such as FLUX.1-dev~\cite{flux1dev2024}, however, use a deterministic ODE for sampling:
\begin{equation}
\mathrm{d}x_t = v_\theta(x_t, t)\,\mathrm{d}t,
\end{equation}
where $v_\theta$ is the learned velocity field. A deterministic solver produces a one-to-one mapping between noise and output, so branching from a shared intermediate state yields identical children. This makes guided search methods inapplicable without modification.

Following Liu \etal~\cite{liu2025flow} and the stochastic interpolant framework~\cite{albergo2025stoas, song2020score}, we convert the probability-flow ODE into an equivalent SDE that preserves the marginal distribution $p_t(x)$ at all timesteps while introducing the stochasticity needed for branching. The key construction proceeds as follows.

The marginal density $p_t(x)$ of the ODE evolves according to the continuity equation
\begin{equation}
\partial_t p_t(x) = -\nabla \cdot [v_t(x)\, p_t(x)].
\end{equation}
An SDE of the form $\mathrm{d}x_t = f_{\mathrm{SDE}}(x_t, t)\,\mathrm{d}t + \sigma_t\,\mathrm{d}w$ produces a marginal density governed by the Fokker--Planck equation
\begin{equation}
\partial_t p_t(x) = -\nabla \cdot [f_{\mathrm{SDE}}\, p_t(x)] + \tfrac{1}{2}\sigma_t^2\,\nabla^2 p_t(x).
\end{equation}
Matching the two expressions yields the required SDE drift:
\begin{equation}
f_{\mathrm{SDE}}(x_t, t) = v_t(x_t) + \tfrac{\sigma_t^2}{2}\,\nabla \log p_t(x_t).
\end{equation}
The corresponding reverse-time SDE~\cite{song2020score} is
\begin{equation}
\mathrm{d}x_t = \bigl[v_t(x_t) - \tfrac{\sigma_t^2}{2}\,\nabla \log p_t(x_t)\bigr]\,\mathrm{d}t + \sigma_t\,\mathrm{d}w.
\label{eq:reverse_sde_general}
\end{equation}
For rectified flow~\cite{liu2022flow} with $x_t = (1-t)\,x_0 + t\,x_1$, the score can be expressed in terms of the velocity field as
\begin{equation}
\nabla \log p_t(x) = -\frac{1}{t}\bigl[x + (1-t)\,v_t(x)\bigr].
\end{equation}
Substituting into \cref{eq:reverse_sde_general} and applying Euler--Maruyama discretization gives the update rule we use in practice:
\begin{equation}
x_{t+\Delta t} = x_t + \Bigl[v_\theta(x_t, t) + \frac{\sigma_t^2}{2t}\bigl(x_t + (1-t)\,v_\theta(x_t, t)\bigr)\Bigr]\Delta t + \sigma_t\sqrt{\Delta t}\;\epsilon, \quad \epsilon \sim \mathcal{N}(0, I).
\label{eq:sde_update}
\end{equation}
The noise schedule $\sigma_t = a\sqrt{t/(1-t)}$ is controlled by a scalar hyperparameter $a$ that governs the level of stochasticity. Setting $a = 0$ recovers the original deterministic ODE, while larger values of $a$ increase diversity at the cost of sample quality. In our experiments, we set $a = 0.3$, which provides sufficient stochasticity for trajectory branching while preserving image fidelity.

We emphasize that this conversion is applied \emph{only} for the guided search baselines (BFS, DFS, ZOS) that require branching at intermediate steps. Our proposed \methodname does not require stochastic sampling, since it generates independent draft candidates from different initial noise seeds and does not branch from shared intermediate states.

\subsection{Evaluation Protocol Details \& Hyper-parameters}
\label{appendix_subsec:eval_protocol_details}

\paragraph{Benchmark subsampling.}
\looseness=-1
GenAI-Bench contains 1,600 prompts and UniGenBench contains 600 prompts. Running all baselines for up to 300\,s per prompt on the full benchmarks would require thousands of GPU-hours per model, making exhaustive evaluation intractable. We therefore uniformly sample 300 prompts from the evaluation set of each benchmark and use these fixed subsets across all experiments. GenEval contains 553 prompts and is evaluated in full.

\paragraph{Evaluation protocol.}
GenEval~\cite{ghosh2023geneval} evaluates compositional text-to-image generation by assessing whether generated images satisfy structured compositional constraints (object presence, counting, spatial relations, attribute binding, etc.). For GenAI-Bench~\cite{li2024genai}, we use VQAScore as the evaluation metric, following the official benchmark release. VQAScore measures text-image alignment by computing the probability that a visual question answering model would answer ``yes'' to the question ``Does this image match the description: \{prompt\}?'' For UniGenBench~\cite{wang2025pref}, we use the officially released composite evaluation metric that combines multiple assessment dimensions including semantic alignment, attribute accuracy, and spatial correctness.

We follow the standard evaluation protocol of generating 4 images per prompt and reporting the average benchmark-specific score across these 4 images. In the inference-time scaling setting, each method generates a pool of candidates under a fixed wall-clock budget and selects the best ones via its respective selection mechanism. We report the average evaluation score over the top-4 images as ranked by the verifier for each method at each time budget.

\paragraph{Hyper-params for different techniques:}
All methods run on NVIDIA H200 GPUs with a batch size of 8: for BoN and \methodname, 8 candidates are explored in parallel per batch; for particle-based methods (BFS, DFS, ZOS), the particle group size is 8. We reserve two GPUs per node, one for generation and one for verification, ensuring that verifier inference does not interfere with generation wall-clock measurements. For BFS we follow Zhang \etal~\cite{zhang2025inferencetimescalingdiffusionmodels} and use their improved configuration: an increasing tempering schedule that upweights later (less noisy) timesteps, Max scoring that tracks the best verifier estimate along each particle's trajectory, and Srinivasan sampling process (SSP) resampling for low-variance particle allocation. For DFS, we use a backtracking budget of $K{=}3$ and backtrack step size $\Delta T{=}10$. For ZOS~\cite{Ma_2025_CVPR}, we use guidance weight $\lambda{=}0.1$. For \methodname, the optimal draft configurations $\phi^*$ obtained via discrete optimization are as follows. Wan2.1 1.3B: $S'{=}49$, $n_{\text{skip}}{=}3$, $\mathcal{L}_{\text{skip}}{=}\{22,23\}$, $f_{\text{full}}{=}5$. Wan2.1 14B: $S'{=}49$, $n_{\text{skip}}{=}2$, $\mathcal{L}_{\text{skip}}{=}\{34,35\}$, $f_{\text{full}}{=}4$. FLUX.1-dev: $S'{=}48$, $n_{\text{skip}}{=}11$, $\mathcal{L}_{\text{skip}}{=}\{40,...,44\}$, $f_{\text{full}}{=}3$.

\subsection{Combining Drafts with Existing Scaling Methods: Additional Details}
\label{appendix_subsec:composability_details}

We use the same draft configuration $\phi^*$ as standalone \methodname. The critique and revision steps both use Qwen2.5-VL-7B~\cite{bai2025qwen25vltechnicalreport}, with the exact prompting templates from the Reflection-Tuning paper~\cite{Zhuo_2025_ICCV}. The reflection loop runs on draft images until the VLM no longer suggests revisions or a maximum iteration count is reached; a full-quality generation is produced only after convergence. Similarly, we apply the same $\phi^*$ (layer skipping and activation proxies) to accelerate denoising steps between BFS verification points, reverting to full computation only after the final pruning decision. All BFS hyperparameters (Increase tempering, Max scoring, SSP resampling) remain identical to the standalone BFS baseline described in Appendix~\ref{appendix_sec:addn_experiment_setup}.

\subsection{RL Post-Training: Additional Details.}
\label{appendix_subsec:grpo_details}

Both the baseline Flow-GRPO and \methodshort-Flow-GRPO share the same core hyperparameters: LoRA $\alpha{=}64$, $r{=}32$, learning rate $10^{-4}$, GRPO clip range $10^{-3}$, noise level $a{=}0.7$, CFG scale 5.0, training denoising steps $T{=}20$, evaluation denoising steps $T{=}50$. Training prompts are drawn from the GenAI-Bench train set combined with the HPSv3 dataset~\cite{ma2508hpsv3}.
The two runs differ only in rollout selection and regularization. In \methodshort-Flow-GRPO, we generate $2G{=}16$ draft rollouts and retain the top-6 and bottom-2 for gradient computation. The top-heavy selection ratio (6 positive, 2 negative) was chosen after sweeping top-heavy, balanced, and bottom-heavy strategies; top-heavy selection empirically performed best, likely because difficult prompts benefit more from having multiple strong positive examples to reinforce than from additional negative signal. The draft configuration $\phi_{\text{grpo}}$ is intentionally less aggressive than the inference-time $\phi^*$: we use activation proxies only ($f_{\text{full}}{=}5$), with no layer skipping or early stopping. This conservative choice reflects a stricter requirement than at inference time: drafts must not only preserve verifier ranking fidelity, but also ensure that reference-model log-probabilities computed from full forward passes on the selected trajectories remain well-calibrated. For the same reason, we increase the KL penalty to $\beta{=}0.1$ (vs.\ $0.05$ for the baseline) to prevent the policy from diverging too far under the draft-selected distribution.

To ensure a fair comparison, the total cost of the expanded rollout pipeline (draft generation + reward scoring + filtering) is calibrated to be approximately equal to, and in practice slightly faster than, the baseline's rollout sampling time. This means any improvement from \methodshort-Flow-GRPO is attributable to better rollout selection,
with no increase in sampling time.
After selection, the GRPO gradient step is identical between the two methods: full forward passes are computed on the same number of trajectories ($G{=}8$), and the policy update uses the same loss and clipping.
Fig.~\ref{fig:grpo_rollout} illustrates the benefit of the expanded rollout pool. For compositionally challenging prompts, the baseline's $G{=}8$ random rollouts often produce visually similar candidates that all fail in the same way, providing limited training signal. The expanded $2G{=}16$ draft pool is more likely to contain at least one rollout that satisfies the prompt, giving the optimizer a concrete positive example to reinforce.

\section{Additional Experiments}

\subsection{Wall-Clock vs.\ NFE on Additional Benchmarks}
\label{appendix_subsec:time_vs_nfe}

\begin{figure}[!h]
    \centering
    \includegraphics[width=0.9\linewidth]{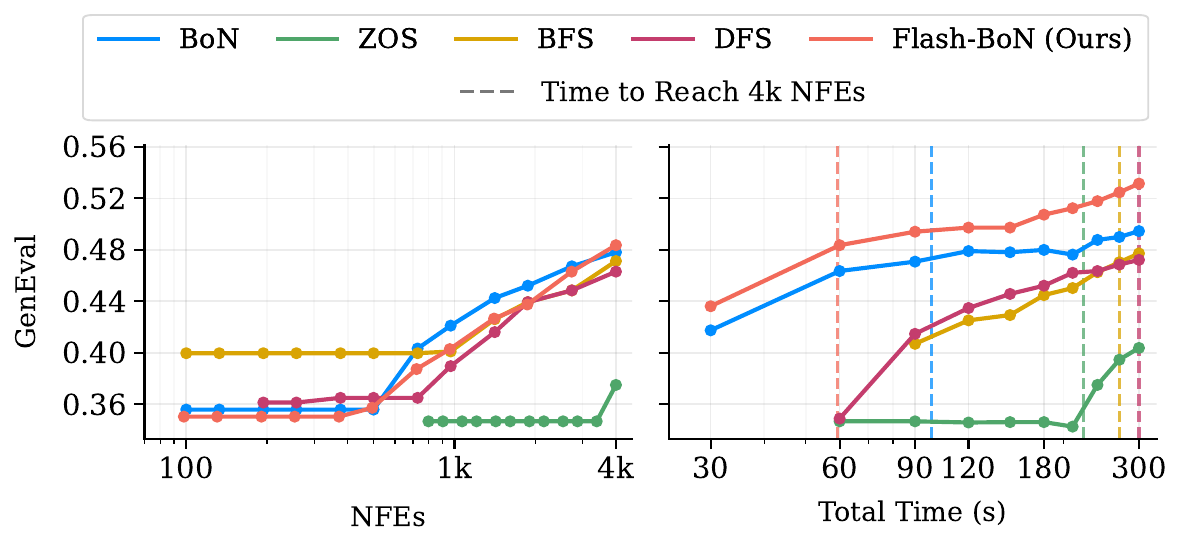}
    \caption{[\textbf{T2I Model:} Wan2.1 1.3B, \textbf{Dataset:} GenEval] Comparison under matched NFE and matched wall-clock budgets. \textbf{Left:} At a fixed NFE budget, Breadth-First Search (BFS) is marginally best at high NFEs. \textbf{Right:} Under equal runtime, the trend reverses. Dashed vertical lines mark the wall-clock time each method needs to reach 4k NFEs (the maximum NFE shown on the left). Methods with frequent verifier calls take longer to reach the same NFE budget, reducing exploration within a fixed runtime. Consequently, BoN-style approaches outperform under realistic latency constraints.}
    \label{fig:geneval_time_wall}
\end{figure}

\begin{figure}[!h]
    \centering
    \includegraphics[width=0.9\linewidth]{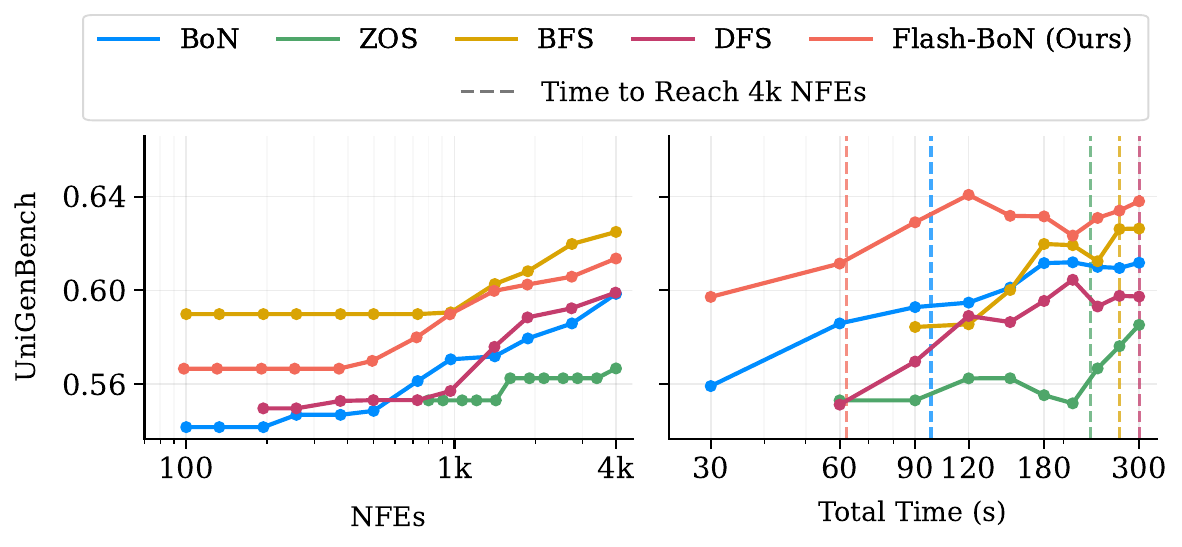}
    \caption{[\textbf{T2I Model:} Wan2.1 1.3B, \textbf{Dataset:} UniGenBench] Comparison under matched NFE and matched wall-clock budgets. \textbf{Left:} At a fixed NFE budget, Breadth-First Search (BFS) is marginally best at high NFEs. \textbf{Right:} Under equal runtime, the trend reverses. Dashed vertical lines mark the wall-clock time each method needs to reach 4k NFEs (the maximum NFE shown on the left). Methods with frequent verifier calls take longer to reach the same NFE budget, reducing exploration within a fixed runtime. Consequently, BoN-style approaches outperform under realistic latency constraints.}
    \label{fig:unigenbench_time_wall}
\end{figure}

\subsection{Reward Score Trajectories}
\label{appendix_subsec:score_trajectories}

\begin{figure}[t]
    \centering
    \begin{minipage}[t]{0.48\linewidth}
        \centering
        \includegraphics[width=\linewidth]{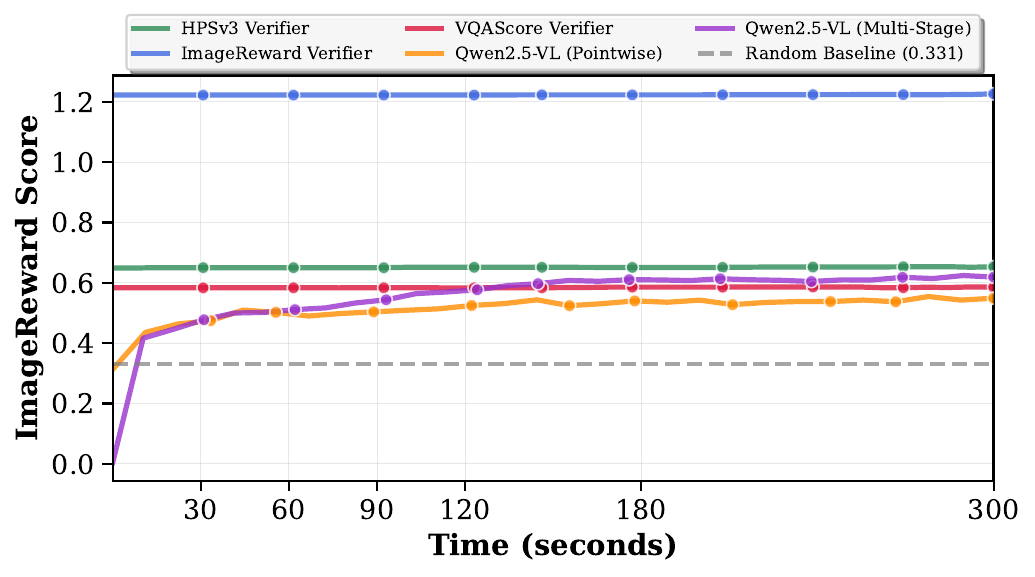}
        \captionof{figure}{\textbf{ImageReward score trajectories under different verifiers (Wan 2.1 1.3B).} Each curve shows the average ImageReward score of the best candidate selected by a given verifier as the wall-clock budget increases. The ImageReward verifier (blue) saturates almost immediately to ${\sim}1.22$, while others plateau much lower. Dashed line: no-scaling baseline (0.331).}
        \label{fig:score_trajectories}
    \end{minipage}%
    \hfill
    \begin{minipage}[t]{0.48\linewidth}
        \centering
        \includegraphics[width=\linewidth]{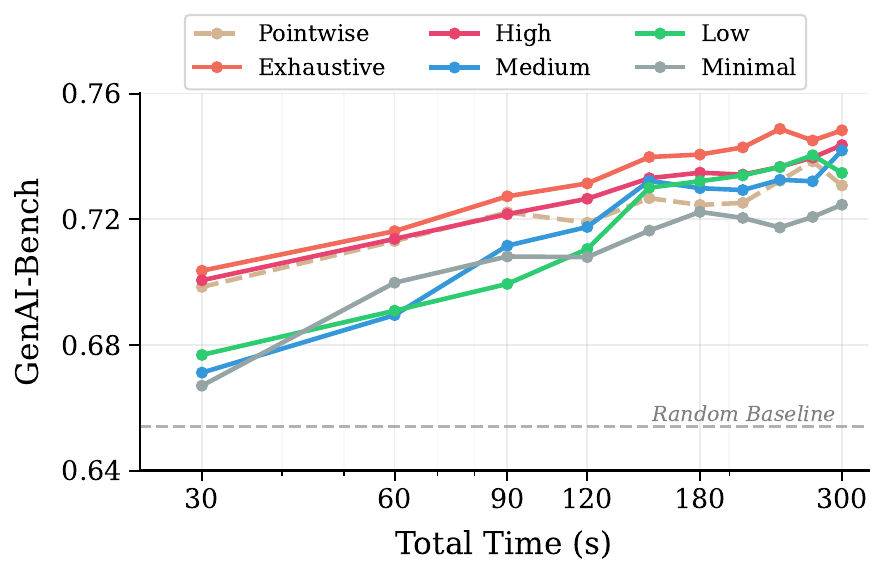}
        \caption{\textbf{Effect of verification budget on selection quality (Wan 2.1 1.3B, GenAI-Bench).} Higher pairwise budgets consistently improve performance over pointwise-only scoring (dashed), with the gap widening as the candidate pool grows at larger time budgets.}
        \label{fig:verification_budget}
    \end{minipage}
\end{figure}

Fig.~\ref{fig:score_trajectories} provides direct evidence for the reward over-optimization hypothesis discussed in \S\ref{sec:experiments}. When ImageReward is used as the verifier, its score jumps from the no-scaling baseline of 0.331 to approximately 1.22 within the first batch of candidates and remains effectively flat for the remaining 300\,s budget. This rapid saturation suggests that ImageReward identifies a narrow band of preferred images early and assigns near-identical scores to all subsequent candidates, providing almost no discriminative signal for further selection.
In contrast, HPSv3 and VQAScore plateau at substantially lower ImageReward scores (${\sim}0.65$ and ${\sim}0.59$), indicating that they select for different image properties that do not maximally exploit ImageReward's scoring criteria. The Qwen-based verifiers (both pointwise and multi-stage) show a gradual upward trend throughout the budget, suggesting more continuous discrimination among candidates. This pattern is consistent with the heatmap in Fig.~\ref{fig:verifier_heatmap}. We also qualitatively observe this effect in Fig.~\ref{fig:verifier_evaluator_bias}: images selected by ImageReward as verifier tend to converge on a narrow aesthetic style regardless of prompt content, whereas selections by other cross-verifiers exhibit greater visual diversity and prompt fidelity.

\subsection{Verification and Selection Ablations}
\label{appendix_subsec:pairwise_vs_pointwise}

\subsubsection{Effect of Verification Budget on Multi-Stage Filtering}
\label{appendix_subsec:verification_budget}

\looseness=-1
Our main experiments use the \texttt{exhaustive} budget for multi-stage filtering. To understand whether scaling verification compute improves selection quality, we sweep five budget levels that control the retention ratios $\rho_1, \rho_2$ and per-candidate comparison count $\kappa$ in our three-stage pipeline (\S\ref{stages}). We evaluate on 100 randomly sampled prompts from GenAI-Bench, as running all budget configurations exhaustively on the full set is not tractable.
All levels perform pointwise scoring on every candidate; they differ only in how many additional pairwise comparisons are allocated per surviving candidate.
\textbf{Minimal}: ${\sim}1$ pairwise comparison per surviving candidate; barely augments pointwise signal.
\textbf{Low}: ${\sim}1.5$ pairwise comparisons per survivor; slight improvement in top-rank discrimination.
\textbf{Medium}: ${\sim}2$ pairwise comparisons per survivor; moderate refinement of the filtered subset.
\textbf{High}: ${\sim}3$ pairwise comparisons per survivor; strong ranking with diminishing returns over medium.
\textbf{Exhaustive}: full pairwise among all stage-3 survivors ($\binom{\rho_1 \rho_2 M}{2}$ comparisons in the final stage); used for all main paper results.

Fig.~\ref{fig:verification_budget} shows GenAI-Bench performance on Wan2.1 1.3B as a function of wall-clock time for each budget level. The trend is intuitive: increasing the pairwise budget improves selection quality, with exhaustive and high budgets consistently outperforming pointwise-only scoring across the full time range. At lower budgets (minimal, low), the sparse pairwise signal provides limited benefit over pointwise alone, and performance tracks closely with the pointwise baseline. These results confirm that scaling along the verification axis yields meaningful gains, complementary to scaling along the exploration axis via cheaper drafts.

\begin{figure}[t]
    \centering
    \begin{minipage}[t]{0.48\linewidth}
        \centering
        \includegraphics[width=\linewidth]{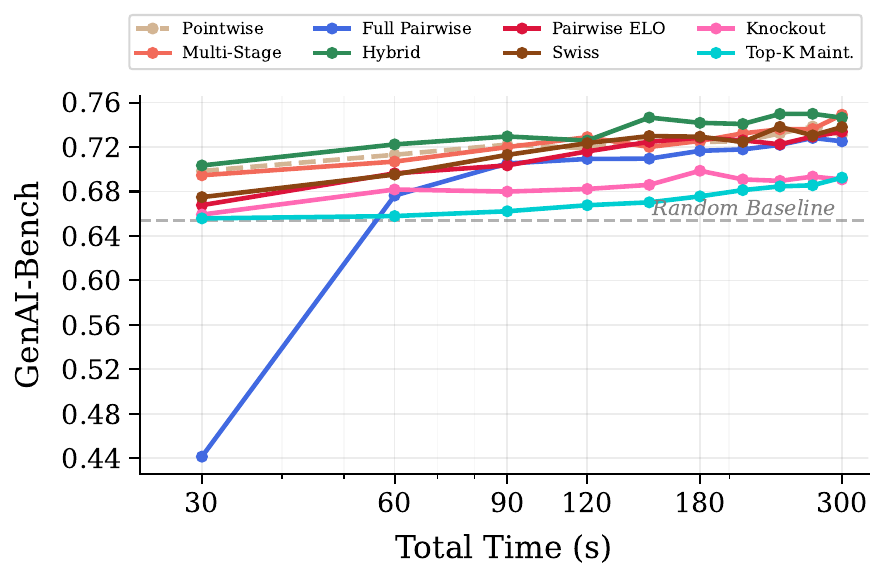}
        \caption{\textbf{Comparison of pairwise ranking strategies (Wan 2.1 1.3B, GenAI-Bench).} Strategies that combine pointwise screening with pairwise refinement (Hybrid, Multi-Stage) consistently lead. Full Pairwise is hampered at small budgets by its quadratic cost, while Knockout and Top-$k$ Maintenance lack sufficient comparison structure for reliable selection.}
        \label{fig:pairwise_strategies}
    \end{minipage}%
    \hfill
    \begin{minipage}[t]{0.48\linewidth}
        \centering
        \includegraphics[width=\linewidth]{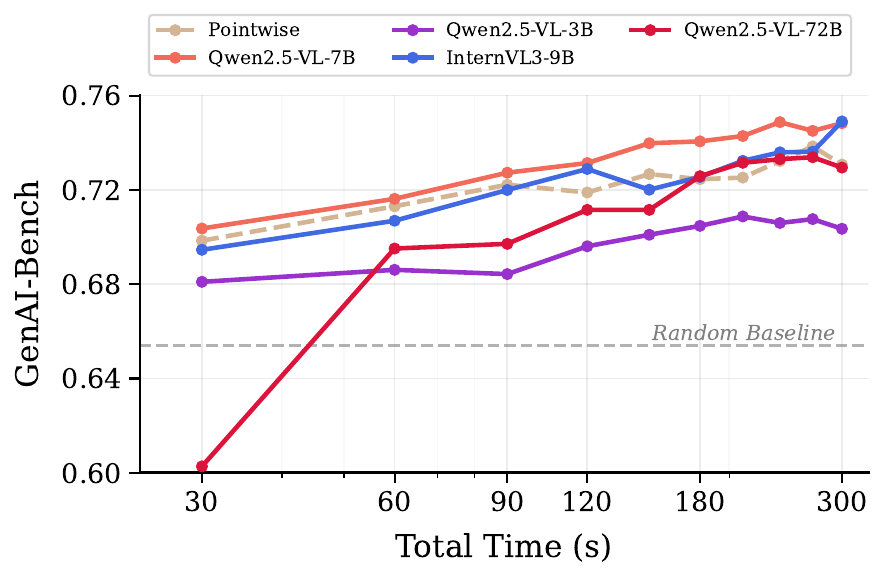}
        \caption{\textbf{Effect of verifier model on selection quality (Wan 2.1 1.3B, GenAI-Bench).} Mid-scale verifiers (Qwen2.5-VL-7B, InternVL3-9B) perform best under wall-clock constraints. The 3B model lacks discrimination ability, while the 72B model's higher inference cost limits exploration, offsetting its stronger per-comparison judgments.}
            \label{fig:verifier_model_ablation}
    \end{minipage}
\end{figure}

\subsubsection{Alternative Pairwise Ranking Strategies.}
\label{appendix_subsec:pairwise_strategies}

In the main paper, we report results with pointwise scoring and our multi-stage filtering procedure. To understand how ranking strategy affects final selection quality, we experiment with six alternatives that span the cost--accuracy spectrum. All use the same VLM (Qwen2.5-VL-7B) and candidate pools; they differ only in how pairwise comparisons are allocated. We evaluate on 100 randomly sampled prompts from GenAI-Bench, as running all strategies exhaustively on the full set is not tractable.

\begin{enumerate}[leftmargin=*,noitemsep]
    \item \textbf{Full Pairwise.} Compares every pair of candidates exhaustively ($\binom{M}{2}$ VLM calls per batch) and ranks by ELO rating. This provides the strongest ranking signal but is prohibitively expensive for large candidate pools.

    \item \textbf{Hybrid.} First scores all candidates via pointwise evaluation, then runs full pairwise comparisons among only the top-$k$ (we use $k{=}3$). Balances the speed of pointwise screening with the accuracy of pairwise refinement, though performance depends on whether the true best candidate survives the pointwise filter.

    \item \textbf{Pairwise ELO.} Runs a configurable number of pairwise comparisons (scaled as a multiple of $M$) with adaptive pairing that matches candidates of similar current rating. More flexible than full pairwise but requires tuning the comparison budget.

    \item \textbf{Swiss Tournament.} Pairs candidates by current score over $\lceil \log_2 M \rceil$ rounds, analogous to a Swiss-system chess tournament. Each round uses $M/2$ comparisons, giving $\mathcal{O}(M \log M)$ total cost. Efficient for moderate pool sizes but provides less resolution at the top than dense pairwise methods.

    \item \textbf{Knockout.} Single-elimination bracket tournament requiring exactly $M{-}1$ comparisons. Extremely cheap but fragile: a single incorrect judgment early in the bracket can eliminate the best candidate permanently.

    \item \textbf{Top-$k$ Maintenance.} Maintains a running leaderboard of the best $k$ candidates across batches. New candidates are compared against the current top-$k$; the leaderboard is updated after each batch.
\end{enumerate}

Fig.~\ref{fig:pairwise_strategies} shows results on Wan2.1 1.3B (GenAI-Bench). We find that strategies integrating both pointwise and pairwise signals (Hybrid, Multi-Stage) consistently outperform the rest, confirming that cheap pointwise pruning followed by targeted pairwise refinement is a strong recipe. Full Pairwise starts well below the pointwise baseline at small budgets because its quadratic comparison cost consumes time that could otherwise be spent exploring more candidates; it catches up only at larger budgets where the ranking advantage compensates for reduced exploration. At the other extreme, Knockout and Top-$k$ Maintenance underperform throughout, as their limited comparison structure is too fragile to reliably identify the best candidate from noisy VLM judgments.

\subsubsection{Effect of Verifier Model Choice}
\label{appendix_subsec:verifier_model_ablation}
Our main experiments use Qwen2.5-VL-7B as the verifier for both pointwise scoring and pairwise comparison. To understand how verifier capacity and architecture affect selection quality, we evaluate \methodname with three additional VLMs spanning different model families and scales.

\textbf{Qwen2.5-VL-3B}: smaller variant of our default verifier; tests whether a $2\times$ reduction in verifier size, and the corresponding wall-clock savings, degrades selection quality.
\textbf{Qwen2.5-VL-72B}: largest Qwen variant; tests whether additional verifier capacity translates to better candidate discrimination.
\textbf{InternVL3-9B}~\cite{wang2025internvl3}: comparable scale to our default but a different architecture and training recipe; tests cross-family generalization of the multi-stage pipeline.

All verifiers use identical prompting templates (pointwise rubric and pairwise rubric) and the same multi-stage filtering pipeline with exhaustive budget. Differences in final selection quality therefore reflect verifier discrimination ability rather than pipeline design. We evaluate on 100 randomly sampled prompts from GenAI-Bench, as running all verifier models exhaustively on the full set is not tractable.

Fig.~\ref{fig:verifier_model_ablation} shows results on Wan2.1 1.3B (GenAI-Bench). The 7B-scale models (Qwen2.5-VL-7B and InternVL3-9B) perform best, with Qwen2.5-VL-7B maintaining a consistent lead and InternVL3-9B converging to competitive performance at larger budgets. This suggests that the multi-stage pipeline generalizes across VLM families at this scale. Qwen2.5-VL-3B underperforms throughout, indicating that the smaller model lacks sufficient discrimination ability for reliable pairwise judgments. Qwen2.5-VL-72B presents an interesting trade-off: while it is presumably a stronger judge per comparison, its substantially higher inference cost reduces the number of candidates that can be explored and verified within the same wall-clock budget, resulting in lower overall performance. This reinforces a recurring theme: under fixed wall-clock constraints, verifier cost must be balanced against exploration breadth, and a moderately sized but fast verifier often outperforms a stronger but slower one.

\subsection{Component Ablations: Contribution of Each Acceleration Knob}
\label{appendix_subsec:component_ablations}

\methodname combines three acceleration knobs---early timestep stopping (TS), layer skipping (LS), and activation proxies (AP)---into a single draft configuration $\phi^*$ (\S\ref{acceleration_knobs}). Here we isolate each knob's contribution and verify that the discrete optimization over $\phi$ is necessary rather than incidental. In~\cref{tab:component_ablation}, starting from a random valid configuration, we progressively enable one knob at a time until reaching the full method.

\begin{table}[t]
\centering
\footnotesize
\caption{\textbf{Component ablation on GenAI-Bench.} Each knob is added progressively on top of the previous. \textbf{AUC}~($\uparrow$): area under the score-vs-wall-clock curve; \textbf{P@90}~($\uparrow$): score at a 90\,s budget; \textbf{Speedup@$P_{\mathrm{BoN}}$(300)}~($\uparrow$)~$=300/\tau$, where $\tau$ is the time to match BoN's 300\,s score ($<$1: never matched). TS / LS / AP $=$ early timestep stopping / layer skipping / activation proxies. Significance for \methodname{} is a one-sided Wilcoxon test vs.\ BoN ($^{*}p{<}0.05$, $^{**}p{<}0.01$, $^{***}p{<}0.001$).}
\label{tab:component_ablation}
\setlength{\tabcolsep}{4pt}
\begin{tabular}{l ccc ccc}
\toprule
 & \multicolumn{3}{c}{\textit{Wan2.1 1.3B}} & \multicolumn{3}{c}{\textit{FLUX.1-dev}} \\
\cmidrule(lr){2-4} \cmidrule(lr){5-7}
Configuration & AUC & P@90 & \shortstack{Speedup@\\$P_{\mathrm{BoN}}$(300)}
              & AUC & P@90 & \shortstack{Speedup@\\$P_{\mathrm{BoN}}$(300)} \\
\midrule
BoN (reference)      & 0.645 & 0.709 & 1.00 & 0.594 & 0.727 & 1.00 \\
\midrule
Random config        & 0.551 & 0.613 & {\scriptsize $<$1} & 0.548 & 0.674 & {\scriptsize $<$1} \\
TS only              & 0.623 & 0.685 & {\scriptsize $<$1} & 0.586 & 0.709 & 1.07 \\
TS\,+\,LS            & 0.646 & 0.719 & 1.43 & 0.621 & 0.731 & 1.24 \\
TS\,+\,LS\,+\,AP     & 0.652 & 0.729 & 2.58 & 0.664 & \textbf{0.757} & 1.76 \\
\midrule
\rowcolor{gray!15}
\methodname{} (full) & \textbf{0.666}\rlap{$^{*}$} & \textbf{0.731}\rlap{$^{*}$} & \textbf{3.33}\rlap{$^{***}$}
                      & \textbf{0.680}\rlap{$^{**}$} & 0.754\rlap{$^{**}$} & \textbf{1.88}\rlap{$^{**}$} \\
\bottomrule
\end{tabular}
\end{table}

\paragraph{Each knob expands the candidate pool.}
A random valid configuration performs well below BoN on both models (AUC 0.551 / 0.548): overly aggressive shortcuts degrade drafts to the point where the verifier can no longer rank them reliably. Enabling early timestep stopping alone recovers most of this gap, and each additional knob yields a further gain in both AUC and iso-quality speedup---layer skipping raises the Wan2.1 speedup to $1.43\times$ and activation proxies to $2.58\times$. The full method, which adds multi-stage pairwise verification on top of all three knobs, attains the best AUC and the largest speedup ($3.33\times$ on Wan2.1 1.3B, $1.88\times$ on FLUX.1-dev). Each knob helps by enlarging the candidate pool explorable within a fixed budget, while pairwise verification keeps selection reliable as that pool grows.

\paragraph{Why discrete optimization is necessary.}
The collapse of the random configuration shows that stacking the three knobs is not enough on its own: the speed--fidelity trade-off must be calibrated per model. The discrete optimization of \S\ref{acceleration_knobs} is a one-time cost of ${\sim}3$ hours per model that generalizes across benchmarks (App.~\ref{appendix_subsec:config_precomputation}) and amortizes over all subsequent inference. Calibration also transfers only partially across models: applying the FLUX.1-dev configuration to Wan2.1 14B yields AUC 0.61---above random and on par with BoN (0.60), but below the 0.68 obtained with native per-model calibration---reinforcing that the optimization is best run once per target model.

\subsection{Few-Step Distilled Models}
\label{appendix_subsec:few_step}

\paragraph{Few-step distilled regime.}
Our main experiments use multi-step T2I models. To test whether the draft-and-select trade-off survives when the denoising budget is already small, we evaluate on FLUX.1-schnell, a 4-step distilled model, with a dedicated configuration ($S'{=}3$, $\mathcal{L}_{\text{skip}}{=}\{40,\ldots,44\}$, $f_{\text{full}}{=}2$). Even with only four steps to work with, \methodname{} outperforms BoN under matched wall-clock budgets (Tab.~\ref{tab:few_step}), reaching BoN's 150\,s score $7.5\times$ faster. The approach is therefore not limited to expensive many-step samplers: as long as a cheaper draft preserves verifier-relevant structure, expanding the candidate pool remains beneficial.

\begin{table}[t]
\centering\footnotesize
\setlength{\tabcolsep}{6pt}
\caption{\textbf{FLUX.1-schnell on GenAI-Bench under wall-clock budgets.} AUC~($\uparrow$); P@20s~($\uparrow$): score at 20\,s; iso-quality speedup to match BoN's 150\,s score. Significance for \methodname{} via one-sided Wilcoxon vs.\ BoN.}
\label{tab:few_step}
\begin{tabular}{l ccc}
\toprule
Method & AUC & P@20s & Speedup@$P_{\mathrm{BoN}}$(150s) \\
\midrule
BoN           & 0.708 & 0.726 & 1.0 \\
ReNO          & 0.689 & 0.720 & {\scriptsize $<$1} \\
\methodname{}  & \textbf{0.733}\rlap{$^{**}$} & \textbf{0.773}\rlap{$^{**}$} & \textbf{7.5}\rlap{$^{***}$} \\
\bottomrule
\end{tabular}
\end{table}

\section{Additional Optimization Details}
\label{appendix_subsec:addn_optimization_details}
\subsection{Configuration for Precomputation}
\label{appendix_subsec:config_precomputation}
Each configuration $\phi$ defines a trade-off between generation speed and
draft quality. To determine an optimal setting $\phi^*$, we formulate a
black-box discrete optimization problem balancing runtime and output fidelity.

\subsubsection{Discrete Optimization}

Our goal is to select a configuration $\phi^*$ that maximizes generation
efficiency without compromising draft quality. We start with a held-out
calibration set $\mathcal{P}$ of 120 prompts, sampled from general-purpose
text-to-image datasets (e.g., Open-Image-Preferences, HPSv2D) and training
splits of standard evaluation benchmarks (e.g., GenEval). We evaluate each
candidate configuration $\phi$ along two axes:

\begin{enumerate}[leftmargin=*, noitemsep]
    \item \textbf{Wall-clock speedup} over the full baseline $\phi_{\text{base}}$,
          measured as $T_{\text{base}} / T(\phi)$.
    \item \textbf{Image fidelity}, measured by perceptual similarity to the
          baseline using LPIPS.
\end{enumerate}

For a given prompt $x$, let $y_{\phi}(x)$ and $y_{\text{base}}(x)$ denote
outputs generated using $\phi$ and the baseline configuration. We define:
\begin{equation}
    \mathrm{Sim}(\phi) = 1 - \frac{1}{|\mathcal{P}|} \sum_{x \in \mathcal{P}}
        \mathrm{LPIPS}(y_{\phi}(x), y_{\text{base}}(x)).
\end{equation}

We then optimize the scalar objective:
\begin{equation}
J(\phi) = \lambda \cdot \left(\frac{T_{\text{base}}}{T(\phi)}\right)
         + (1 - \lambda) \cdot \mathrm{Sim}(\phi),
\end{equation}
where $\lambda \in [0,1]$ controls the speed--quality trade-off.

Because $J(\phi)$ is non-differentiable and expensive to evaluate (requiring
full image generation and LPIPS computation), and the search space for $\phi$
includes mixed discrete and ordinal variables, we use \emph{dual
annealing}~\cite{Tsallis:1987eu, XIANG1997216} as our optimizer. Dual
annealing alternates between global exploration via stochastic proposals and
local refinement in promising regions. At iteration $t$, a new configuration
$\phi'$ is accepted with probability
\begin{equation}
p_{\text{acc}}(t) = \min\left\{1,\; \exp\left(
    \frac{J(\phi') - J(\phi_t)}{\tau_t}\right)\right\},
\end{equation}
where $\tau_t$ is the annealing temperature at step $t$. We run the optimizer
for 200 iterations per model, evaluating all configurations. Optimization
typically completes in 2--4 hours depending on model size.

\subsubsection{Pareto Selection}

\begin{figure}
    \centering
    \includegraphics[width=\linewidth]{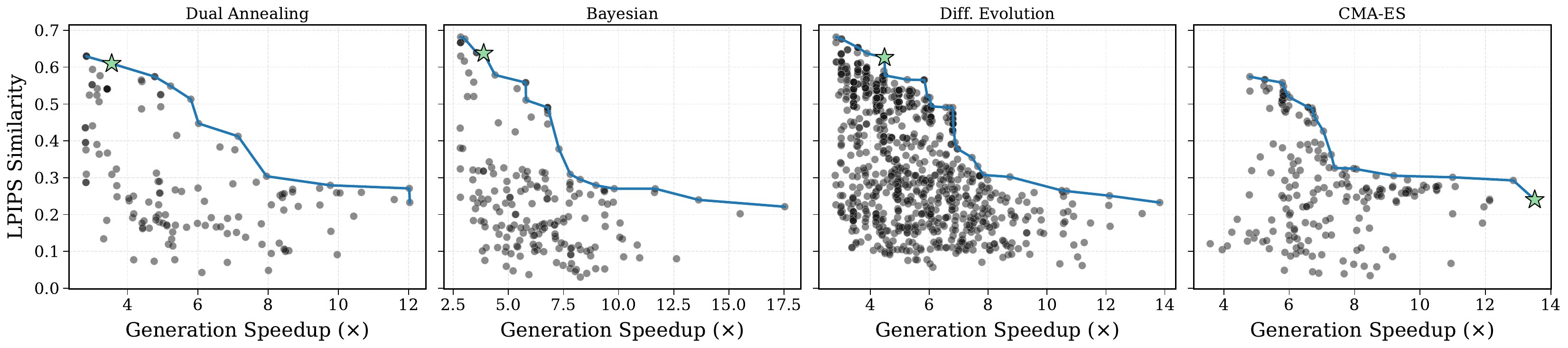}
    \caption{[\textbf{T2I Model:} Wan2.1 1.3B] Pareto frontier of draft configurations showing speedup versus LPIPS similarity to full-compute outputs. The selected configuration $\phi^*$ attains high similarity with substantial speedup across a wide array of discrete optimizers}
    \label{fig:optimizer_grid}
\end{figure}

After evaluating all configurations, we plot the speed--fidelity trade-off and
compute the empirical Pareto frontier, shown in Fig.~\ref{fig:optimizer_grid}.
The scatter highlights a clear trade-off: very large speedups tend to incur
significant fidelity loss, while high-similarity drafts cluster at moderate
acceleration factors. We select $\phi^*$ by maximizing speedup subject to a
minimum similarity constraint, $\mathrm{Sim}(\phi) \ge 0.6$ (star in
Fig.~\ref{fig:optimizer_grid}). The chosen configuration lies in the
high-similarity region of the frontier, providing substantial acceleration
while preserving draft faithfulness for reliable downstream verification
and refinement.

\subsubsection{Effect of Discrete Optimizer}
We compare dual annealing against three alternative black-box optimizers: Bayesian optimization (surrogate-based sequential model), differential evolution (population-based evolutionary search), and CMA-ES (covariance matrix adaptation, which adapts a multivariate Gaussian proposal distribution). Fig.~\ref{fig:optimizer_grid} shows the Pareto frontiers obtained by each method on Wan2.1 1.3B.

Dual annealing, Bayesian optimization, and differential evolution all converge to remarkably similar configurations. Dual annealing: $S'{=}49$, $n_{\text{skip}}{=}3$, $\mathcal{L}_{\text{skip}}{=}\{22,23\}$, $f_{\text{full}}{=}5$. Bayesian: $S'{=}45$, $n_{\text{skip}}{=}0$, $\mathcal{L}_{\text{skip}}{=}\{22,23\}$, $f_{\text{full}}{=}4$. Differential evolution: $S'{=}46$, $n_{\text{skip}}{=}0$, $\mathcal{L}_{\text{skip}}{=}\{22,23\}$, $f_{\text{full}}{=}5$. All three identify the same layer-skip range and similar caching frequencies, suggesting that the objective landscape has a well-defined basin in this region. Minor differences in $S'$ and $n_{\text{skip}}$ reflect optimizer-specific exploration patterns but yield nearly identical Pareto frontiers.
CMA-ES is the exception: its selected configuration ($S'{=}48$, $n_{\text{skip}}{=}5$, $\mathcal{L}_{\text{skip}}{=}\{1,\ldots,27\}$, $f_{\text{full}}{=}7$) skips the majority of layers, which achieves high speedup but at substantial fidelity cost. CMA-ES's Gaussian proposal distribution is better suited to smooth, continuous landscapes and appears to over-explore aggressive layer-skipping regimes from which it does not recover, resulting in a weaker Pareto frontier overall.

\subsubsection{Effect of Calibration Prompts}

Our default calibration set $\mathcal{P}$ draws from general-purpose datasets (Open-Image-Preferences, HPSv3) to ensure the learned configuration is prompt-agnostic. We test whether in-distribution prompts improve optimization by constructing an alternative calibration set of 100 GenAI-Bench prompts from a non-overlapping split. The resulting configuration ($S'{=}42$, $n_{\text{skip}}{=}1$, $\mathcal{L}_{\text{skip}}{=}\{17,18\}$, $f_{\text{full}}{=}5$) is structurally similar to the general-purpose one, with comparable speedup ($3.9\times$) but slightly higher fidelity ($\mathrm{Sim}{=}0.64$ vs.\ $0.60$). This suggests that in-distribution prompts can modestly improve the speed--fidelity trade-off, but the general-purpose calibration set already captures the relevant redundancy structure well enough for robust optimization.

\section{Qualitative Examples}
\vspace{-1em}

\subsection{Correlation between Verifier and Generation Performance}
\label{appendix_subsec:verifier_eval_diff}

\begin{figure}[!t]
    \centering
    \includegraphics[width=\linewidth]{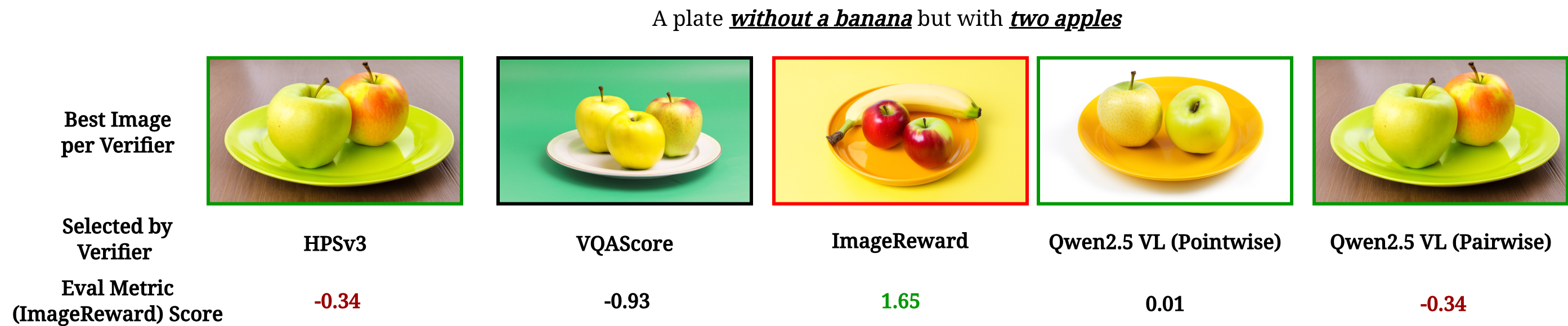}
    \caption{Example of same-metric bias when the verifier and evaluator are both ImageReward. For the prompt ``a plate without a banana but with two apples,'' each verifier selects its top-ranked image, and all selected images are then evaluated using ImageReward. The image selected by the ImageReward verifier contains a banana and therefore violates the prompt, yet it receives the highest ImageReward score. This shows that using the same metric for both selection and evaluation can reward images that exploit evaluator-specific bias rather than genuinely satisfy the prompt.}
    \label{fig:verifier_evaluator_bias}
\end{figure}

The choice of verifier is a crucial factor governing the search process. However, a verifier may be disproportionately weak relative to the generator, for instance when evaluating outputs from a recent model using a much older reward model. In such cases, verifier scores may not meaningfully discriminate between candidates, leading to misleading conclusions. We therefore recommend manually inspecting the alignment between verifier judgments and actual generation quality, at least for a representative subset of samples.

Figure~\ref{fig:verifier_evaluator_bias} illustrates a complementary failure mode that arises when the verifier and the evaluation metric are the same. In this example, each verifier selects its top candidate for the prompt ``a plate without a banana but with two apples,'' and all selected images are then scored by ImageReward. The ImageReward verifier selects an image that clearly violates the prompt because it contains a banana, yet this image also receives by far the highest ImageReward score. This suggests that the search procedure can exploit biases of the evaluator rather than improve true prompt alignment. Put differently, when the verifier and evaluator coincide, the verifier can overfit to the evaluator's preferences, making the resulting scores look artificially strong. For this reason, agreement between verifier and evaluator should be interpreted with caution, especially when the same model is used for both roles.

Conversely, if the generator itself lacks diversity across noise seeds, the search process becomes limited regardless of verifier quality. When most candidates are visually similar, a verifier targeting aesthetics or fidelity has little signal to differentiate among them. It is therefore important that the verifier metric be chosen to account for the diversity characteristics of the generator, ensuring a productive search landscape.

\vspace{-1em}
\subsection{GRPO Expanded Rollout}
\label{appendix_sec:qual_example_grpo}
\begin{figure}[!t]
    \centering
    \includegraphics[width=\linewidth]{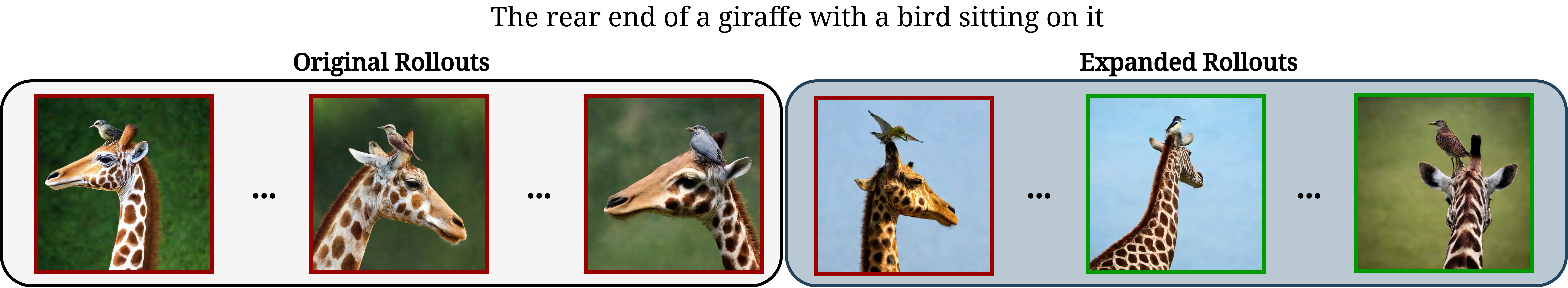}
    \caption{\textbf{Expanded rollouts improve training signal for difficult prompts.} Prompt: ``The rear end of a giraffe with a bird sitting on it,'' the original rollouts (left, red borders) all show the giraffe from the front or side. The expanded draft pool (right) contains candidates with greater diversity, including images that correctly depict the rear view (green borders), providing the optimizer with informative positive examples that would otherwise be absent.}
    \label{fig:grpo_rollout}
\end{figure}

Fig.~\ref{fig:grpo_rollout} illustrates why expanded rollouts may improve training signal. When the baseline's $G{=}8$ random rollouts fail to produce a high signal rollout, the expanded $2G{=}16$ draft pool covers a wider region of the output space, making it substantially more likely to include at least one correct positive example for the optimizer to reinforce.

\subsection{Diversity Qualitative Examples}
\label{appendix_subsec:diversity_qualitative}

\begin{figure}[!h]
    \centering
    \includegraphics[width=\linewidth]{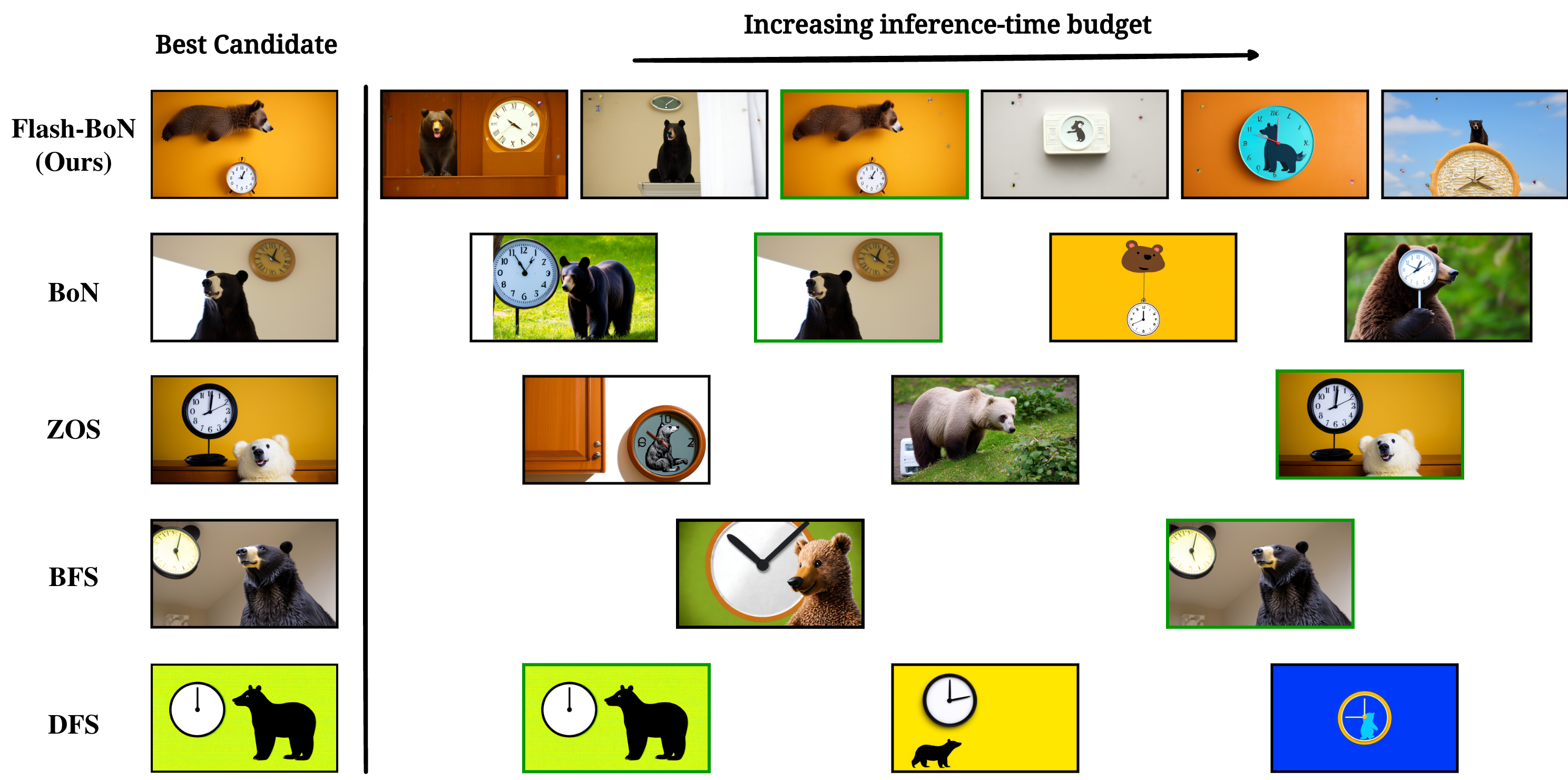}
    \caption{\textbf{Diverse candidate exploration as inference-time budget grows} (prompt: ``a photo of a bear above a clock''). Each row shows the best candidate selected by a method at increasing time budgets, with the leftmost column highlighting the final selection. \methodname explores candidates in the cheap draft space and applies selective refinement only to the best, enabling substantially broader exploration within the same budget. Guided search methods (ZOS, BFS, DFS) generate fewer candidates due to intermediate verifier overhead. The number of images per row is illustrative and does not reflect exact speedup ratios.}
    \label{fig:candidate_progression}
\end{figure}

\begin{figure}
    \centering
    \includegraphics[width=\linewidth]{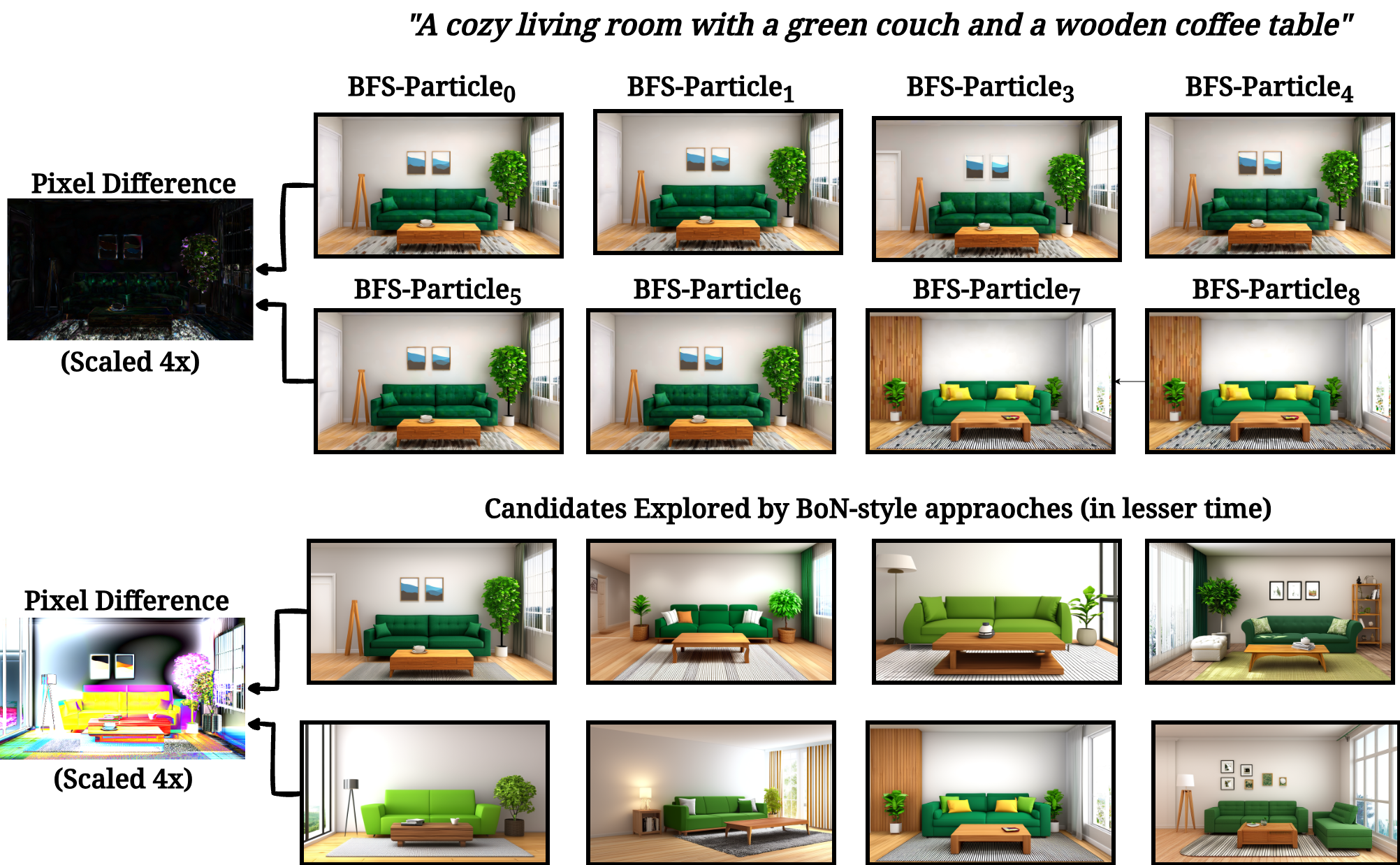}
    \caption{\textbf{BFS particles vs.\ BoN-style candidates.} Top: BFS particles converge to near-identical outputs despite starting from independent seeds; pixel differences (left, scaled $4\times$) are imperceptible. Bottom: BoN-style candidates explore diverse compositions, layouts, and styles within the same prompt.}
    \label{fig:bfs_particle_examples}
\end{figure}

As discussed in \S\ref{sec:experiments}, candidate pool diversity is strongly correlated with final selection quality (Pearson $r{=}0.75$). Fig.~\ref{fig:candidate_progression} illustrates this across time budgets: \methodname explores a visually diverse set of compositions in the cheap draft space before selecting and refining the best candidate, while guided search methods produce fewer and more similar candidates due to intermediate verification overhead.
Fig.~\ref{fig:bfs_particle_examples} highlights the contrast more directly on a single prompt. The BFS particles (top) collapse into near-identical compositions after repeated pruning and resampling: the pixel difference map (scaled $4\times$) confirms that inter-particle variation is confined to imperceptible texture-level differences. This is a direct consequence of local search: upsampling high-scoring trajectories concentrates exploration in a narrow region, and if that region is suboptimal, additional compute cannot escape it. In contrast, BoN-style candidates (bottom) exhibit substantial diversity in layout, lighting, furniture style, and camera angle, as reflected in the much more saturated pixel difference map. This broader coverage is what enables the verifier to identify meaningfully different compositions rather than choosing among near-duplicates.

\section{System Prompts}
\label{appendix_sec:system_prompts}
We provide the system prompts we used during the pairwise and pointwise verification rounds conducted with Qwen2.5-VL-7B (\S\ref{stages}).
\begin{tcolorbox}[colback=gray!10, colframe=gray!50, breakable]
\textbf{System Prompt: Pointwise Scoring}\\[0.5em]
\ttfamily\small
You are a strict prompt-adherence evaluator. Ignore minor visual quality issues.
Judge the image on \textbf{only two criteria}:
\begin{enumerate}
    \item \textit{Object \& Scene Match} - Are \textbf{all} objects, characters, and key scene elements requested in the prompt present?
    \item \textit{Visual Elements Match} - Do colors, styles, compositions, lighting, or other visual attributes appear as described?
\end{enumerate}
Assign a single integer \textit{Score} (0-10) representing overall prompt adherence:
\begin{itemize}
    \item 9-10: Perfect semantic match - ALL elements clearly present and correct
    \item 7-8: Strong match - Most elements correct, minor omissions acceptable
    \item 5-6: Moderate match - Core concept present but significant gaps
    \item 3-4: Weak match - Only partial/vague connection to prompt
    \item 1-2: Minimal match - Barely any connection
    \item 0: Complete failure - No connection whatsoever
\end{itemize}
Finally, output \textbf{exactly} three newline-separated lines \textit{with no extra text}:\\[0.3em]
Reasoning: $\langle$1-2 sentences$\rangle$\\
Verdict: $\langle$yes\textbar no$\rangle$\\
Score: $\langle$0-10$\rangle$
\end{tcolorbox}
\begin{tcolorbox}[colback=gray!10, colframe=gray!50, breakable]
\textbf{System Prompt: Pairwise Scoring}\\[0.5em]
\ttfamily\small
You are comparing two AI-generated images (Image A and Image B) against the same text prompt.
Your task: Determine which image BETTER matches the prompt.\\[0.5em]
Consider these criteria:
\begin{enumerate}
    \item \textit{Object \& Scene Match} - Which image has more of the requested objects/characters/elements?
    \item \textit{Visual Elements Match} - Which image better matches described colors, styles, compositions?
\end{enumerate}
Rules:
\begin{itemize}
    \item Focus ONLY on prompt adherence, not general image quality
    \item If both are equally good/bad, still pick the slightly better one
    \item Only output `TIE' if they are truly indistinguishable
\end{itemize}
Output \textbf{exactly} three lines:\\[0.3em]
Reasoning: $\langle$1-2 sentences comparing the images$\rangle$\\
Winner: $\langle$A, B, or TIE$\rangle$\\
Confidence: $\langle$high, medium, low$\rangle$
\end{tcolorbox}

\end{document}